\documentclass[5p,11pt,nonatbib]{elsarticle}

\usepackage{graphicx}
\usepackage{subcaption}
\usepackage[figuresright]{rotating}
\usepackage{booktabs}
\usepackage{array}
\usepackage{multirow}
\usepackage{tabularx}
\usepackage{xcolor}
\usepackage{amsmath,amssymb}
\usepackage{hyperref}

\usepackage[backend=biber,style=authoryear,maxcitenames=2,maxbibnames=99]{biblatex}
\journal{Journal of Archaeological Science}

\begin{document}

\begin{frontmatter}

\title{Decoding the Past: An Uncertainty-Aware Deep Learning\\ Framework for Sex Attribution in Prehistoric Hand Stencils}

\author[aff1]{Karel Becerra}
\ead{karel.becerra@azyri.com}
\author[aff2]{Boris Mederos}
\ead{boris.mederos@uacj.mx}
\author[aff3]{Dean Snow}
\ead{drs17@psu.edu}
\author[aff4, cor1]{Ram\'on A. Mollineda}
\ead{mollined@uji.es}

\affiliation[aff1]{organization={Data Science and AI Division, Azyri},
city={Miami}, state={Florida}, country={USA}}

\affiliation[aff2]{organization={Departamento de Física y Matemáticas, Instituto de Ingeniería y Tecnología, Universidad Autónoma de Ciudad Juárez}, city={Ciudad Juárez}, country={Mexico}}

\affiliation[aff3]{organization={Professor Emeritus of Anthropology at Pennsylvania State University},
city={Pennsylvania}, country = {USA}}

\affiliation[aff4]{organization={Institute of New Imaging Technologies, Universitat Jaume I},
city={Castelló de la Plana}, country = {Spain}}

\cortext[cor1]{Corresponding author: mollined@uji.es}

\begin{abstract}
%ChatGPT (finely tuned by M)
Determining the biological sex of the individuals who created Upper Paleolithic hand stencils remains a challenging problem due to the absence of ground truth, population differences between contemporary and prehistoric groups, and the uncertainty introduced by image degradation. Traditional morphometric methods suffer from high structural overlap across sexes, poor cross-population generalizability, and subjective feature engineering. This study presents an uncertainty-aware deep learning framework for sex attribution in prehistoric hand stencils that explicitly models, propagates, and aggregates uncertainty throughout the analytical pipeline. The methodology combines dual image processing, dual contour extraction, structured silhouette augmentation, model architectural diversity, and ensemble-based decision aggregation. The pipeline generates twelve plausible silhouette realizations per stencil to capture boundary uncertainties, which are processed by two ensembles of ten deep neural networks each (EfficientNet-B3 and MobileViT-S) trained on 14,036 contemporary hand samples. Furthermore, a triangulated validation scheme integrates ensemble predictions with unsupervised 2D latent-space manifold mapping (UMAP + k-NN) and explainable AI spatial attributions (LayerCAM) to ensure anatomical consistency. On contemporary data, ensemble models achieve strong classification performance, with accuracies exceeding 88\% in older age groups. When applied to prehistoric stencils, the framework produces both sex predictions and confidence measures of internal agreement, enabling the distinction between morphologically stable and ambiguous cases. The convergence observed across ensemble predictions, latent-space structure, and interpretability analyses suggests that uncertainty can be transformed into an explicit and measurable component of archaeological inference, providing a robust, transparent, and reproducible approach to decoding ancient rock art.\\

\end{abstract}


\end{frontmatter}

\section{Introduction}\label{sec:intro}

Echoes of ancient hands in cave art can be found worldwide, with hand stencils—negative handprints created by spraying pigment around the hand—being the most common form \parencite{snow:2006,lundborg:2014}. They provide some of the earliest evidence of symbolic behavior and artistic expression, that some researchers have related to personal signatures, ritual marks, or protective symbols to ward off danger \parencite{janssens:1957,permana:2017}. As suggested by \parencite{fernandez:2025a}, wide\-spread participation in rock art may indicate group cohesion, as well as an intention %to forge a collective identity and 
to transmit cultural values across generations. 

Recent research has leveraged advanced, non-des\-truc\-tive techniques for analyzing and measuring archaeological hands \parencite{snow:2013}. 
To mitigate the inaccuracies and optical distortions associated with 2D photographs and in-situ measurements, photogrammetric 3D models of archaeological hand stencils have been used to generate 2D orthophotos, on which landmarks and semilandmarks were subsequently identified for morphometric analysis \parencite{fernandez:2022, fernandez:2025a}. 
%2D orthoimages have been generated from high-resolution 3D models to correct for distortions caused by perspective and surface relief of cave walls \parencite{fernandez:2022}. 
While most studies have focused on estimating the sex of the artists behind Palaeolithic cave art \parencite{wang:2010,snow:2013}, recent groundbreaking work has expanded this focus to include age prediction \parencite{fernandez:2025a}. Strong evidence suggests that both sexes, including children and subadults, were actively involved in these artistic practices, challenging the long-standing belief that they were primarily created by adult males.

The biological evidence for sexual dimorphism in human hands is extensive and well-established, encompassing differences in hand size, shape, bone structure, dermatoglyphic patterns, and morphometric traits \parencite{snow:2013,karakostis:2015,bindurani:2016,yune:2019,polcerova:2023}. %,fernandez:2024b}.
Studies consistently show that males have larger and more robust hands—with broader palms and thicker finger bones—than females, with these differences emerging in early childhood (by age 3) and becoming more pronounced during development \parencite{karakostis:2015,bindurani:2016}. %,fernandez:2024b}. 
Skeletal analyses reveal that the proximal phalanges are significantly larger in men, with the greatest dimorphism observed in the thumb \parencite{karakostis:2015}. Hand proportions, such as the hand index (breadth-to-length ratio), are also reliably higher in males and have proven useful for sex estimation across populations, albeit with moderate accuracy \parencite{bindurani:2016,suleiman:2024}. Dermatoglyphic studies demonstrate consistent sex-based differences in ridge counts, particularly on the radial sides of the thumb and index finger, which may reflect prenatal developmental influences \parencite{polcerova:2023}. Geometric morphometric analyses and recent advances in machine learning further support the presence of subtle but detectable shape and proportion differences that can be leveraged for sex classification, even in cases where such differences are imperceptible to human observers \parencite{kralik:2014,yune:2019,fernandez:2024b}. Collectively, these findings provide robust multidisciplinary support for the presence of sexual dimorphism in human hands. %, both in living populations and archaeological contexts.

The analysis of prehistoric handprints to determine the sex of Paleolithic cave artists has attracted growing interest in archaeology, offering insights into gender roles. However, while some studies have shown promising results, critical methodological and po\-pu\-la\-tion-based challenges remain. Traditional morphometric methods for sexing prehistoric handprints, based on size, angles, and digit ratios, are limited by significant hand size overlap between sexes and ages, poor generalizability across populations, and low accuracy (often around 60–65\%) \parencite{snow:2006,snow:2013,jowaheer:2011, galeta:2014}. Manning index (2D:4D ratio) also show substantial within- and between-population variation \parencite{mcintyre:2006,manning:2007}, reducing reliability unless population parameters are known \parencite{nelson:2006}. Another challenge, noted by \parencite{gunn:2006}, is that stencil images created from a same hand can vary in size measurements by up to 5 mm. Unsurprisingly, earlier studies aimed at estimating the sex of prehistoric hand stencils have yielded notably conflicting results 
%, leading to divergent interpretations regarding the original handprints' meaning or purpose 
\parencite{nelson:2017}. %Accordingly, morphometric methods may have limited reliability when applied to early human populations, as the use of modern standards in interpreting prehistoric communities could introduce potential biases \parencite{galeta:2014}. %,fernandez:2024}.

%Traditional approaches relied on manually extracted features, such as geometric measurements or digit ratios, which suffer from severe limitations. First, handcrafted features (e.g., hand size and finger ratios) are heavily based on domain expertise, which is subjective, prone to human bias and can limit performance: manually chosen features may not optimally represent the data’s underlying structure. Second, they often fail to generalize across diverse datasets, as they may not account for cross-dataset variations. Third, these features tend to capture only predefined, low-level traits (e.g., sizes, angles, ratios), missing subtle high-level patterns that can be more discriminating for classification tasks. Additionally, manual feature extraction struggles with scalability—each new task or dataset may require re-engineering features.

Traditional approaches rely on manually extracted handcrafted features, such as geometric measurements and digit ratios, which are heavily dependent on subjective domain expertise, often fail to generalize across datasets, capture only predefined low-level traits, and scale poorly. %, as new tasks typically require redesigning features.
Some of the aforementioned limitations were addressed by an early automated approach \parencite{wang:2010} combining image processing techniques with a machine learning algorithm (Support Vector Machines) to extract a number of hand features.  
%Rather than relying on manual measurements, the method extracted 33 engineered features based on the hand contour, providing a more objective and efficient means of studying sexual dimorphism. %Discriminant functions trained on a limited number of scanned images of modern hands were then applied to ancient handprints. 
A more recent work \parencite{fernandez:2025a} also automates the location of landmarks, the calculation of engineered features, and decision-making using discriminant analysis. Nonetheless, subjectivity persists in the selection of human-designed features, while %and the discriminant potential of 
more abstract morphological patterns remains unexplored. 

Recent work \parencite{mollineda:2025,fernandez:2025b} introduces deep learning techniques for sex prediction from prehistoric handprints. In the absence of annotated ancient data, deep neural networks (DNNs) are trained on contemporary hand silhouettes and then transferred to Paleolithic samples. By learning hierarchical representations directly from images, these models avoid manual feature engineering and capture complex, data-driven patterns that may enhance accuracy and robustness across domains. The underlying hypothesis is that hand morphology encodes high-level, sex-informative features that remain sufficiently invariant across age and long-term evolutionary change.

\begin{figure*}[t]
	\centering
	% First Image
	\begin{subfigure}[c]{0.24\textwidth}
		\centering
		\includegraphics[width=\textwidth]{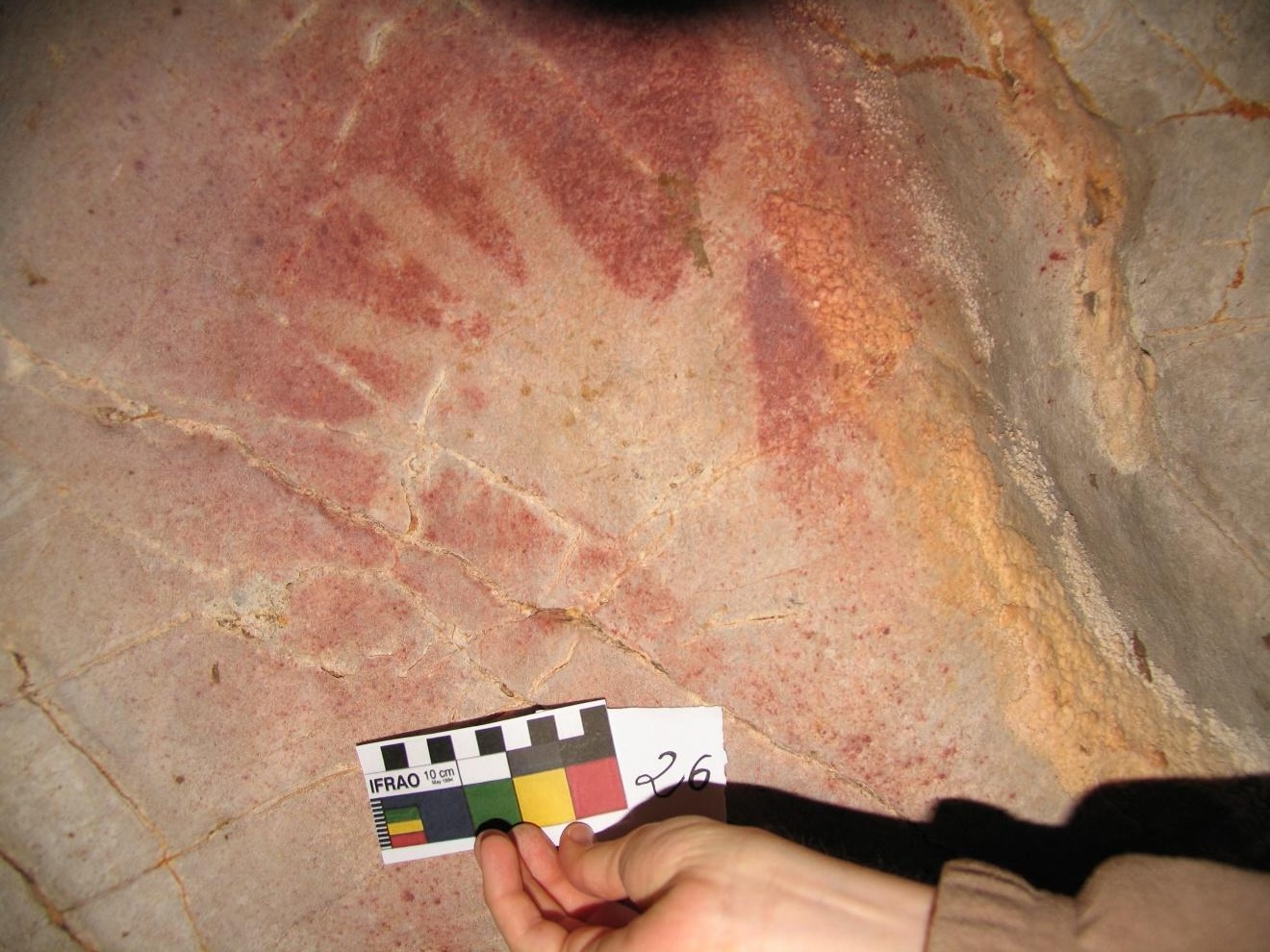}
		\caption{img\_26}
		\label{fig:img26}
	\end{subfigure}
	\hfill
	% Second Image
	\begin{subfigure}[c]{0.24\textwidth}
		\centering
		\includegraphics[width=\textwidth]{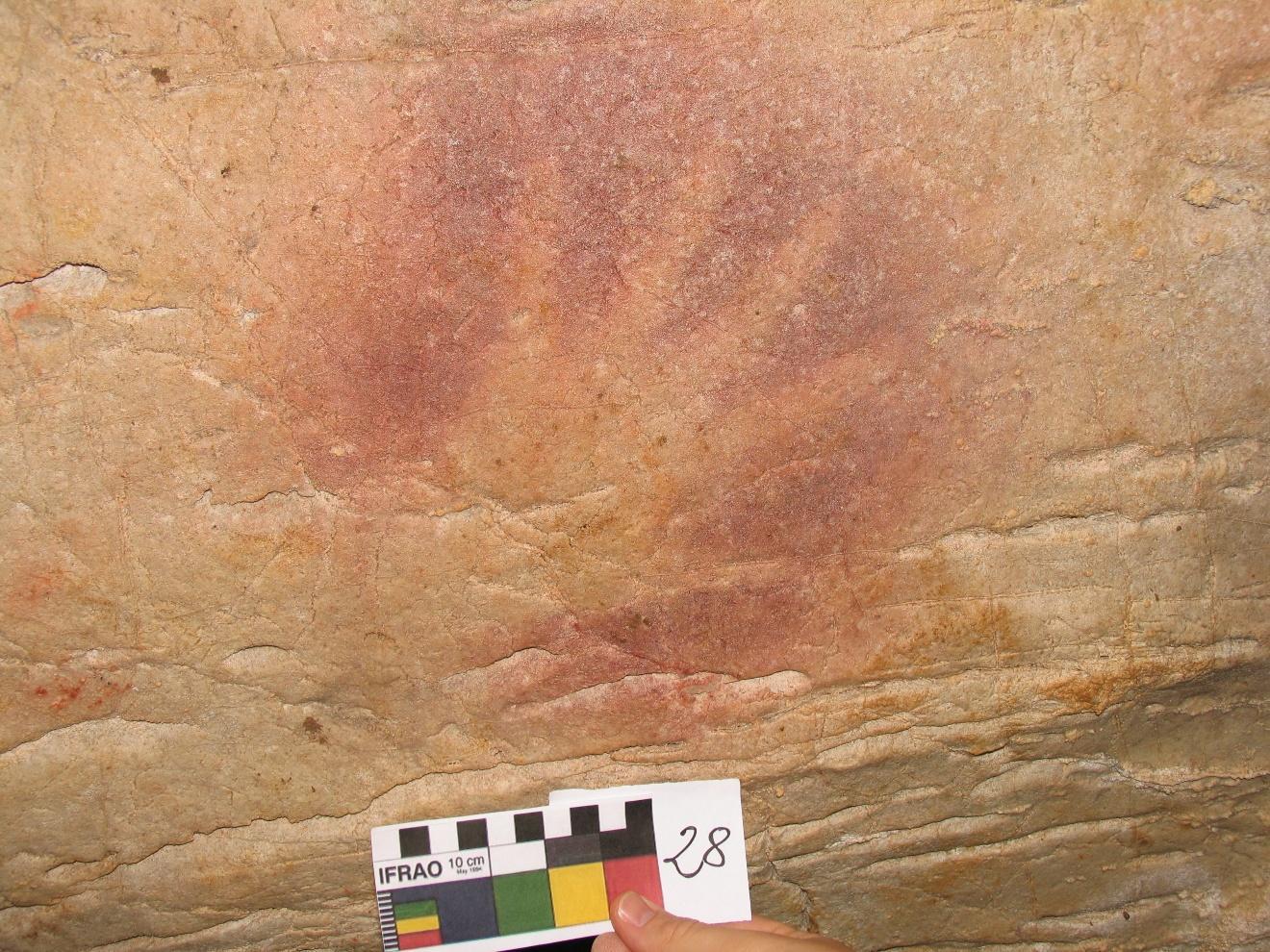}
		\caption{img\_28}
		\label{fig:img28}
	\end{subfigure}
	\hfill
	% Third Image
	\begin{subfigure}[c]{0.24\textwidth}
		\centering
		\includegraphics[width=\textwidth]{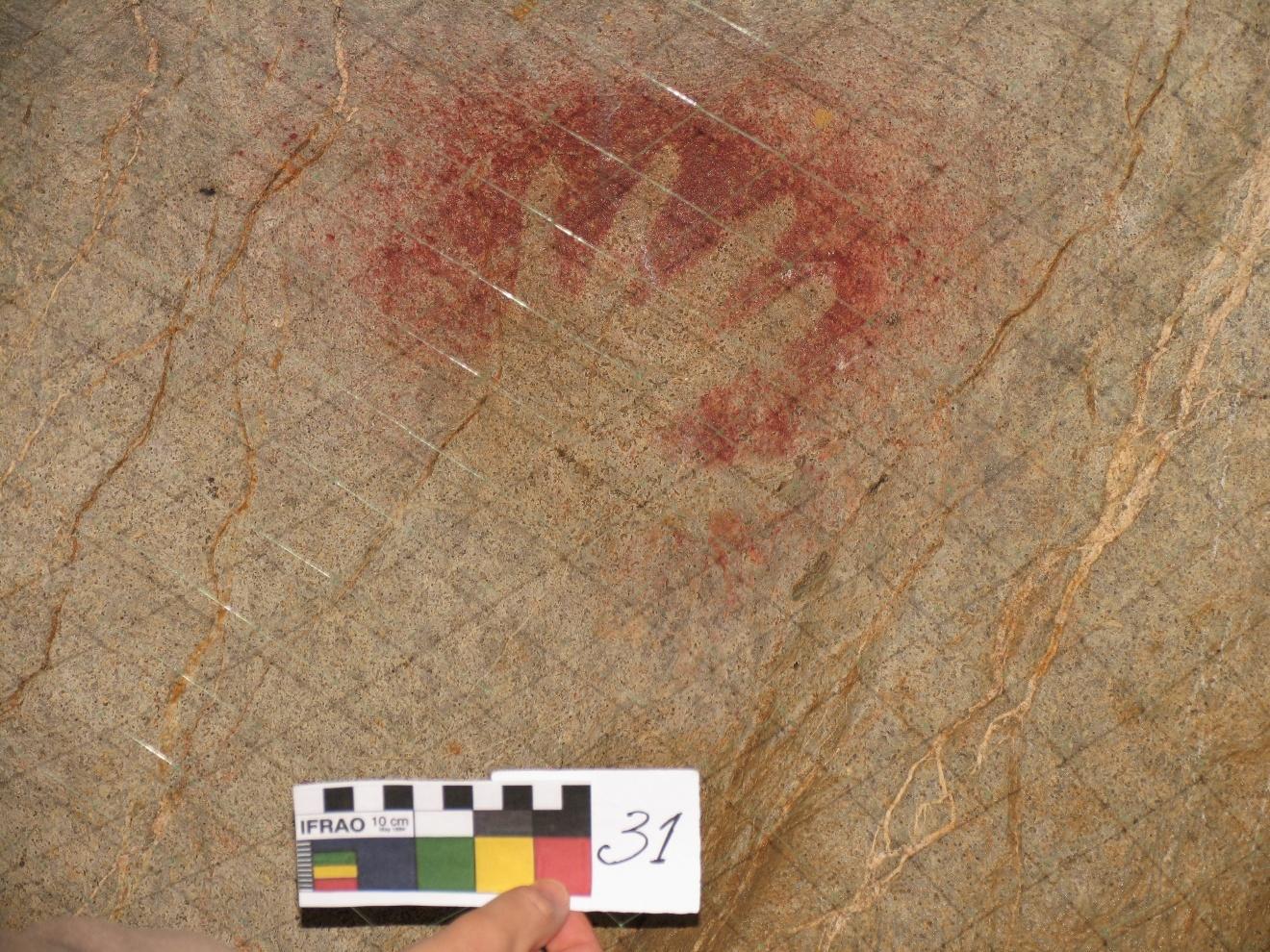}
		\caption{img\_31}
		\label{fig:img31}
	\end{subfigure}
	\hfill
	% Fourth Image
	\begin{subfigure}[c]{0.24\textwidth}
		\centering
		\includegraphics[width=\textwidth]{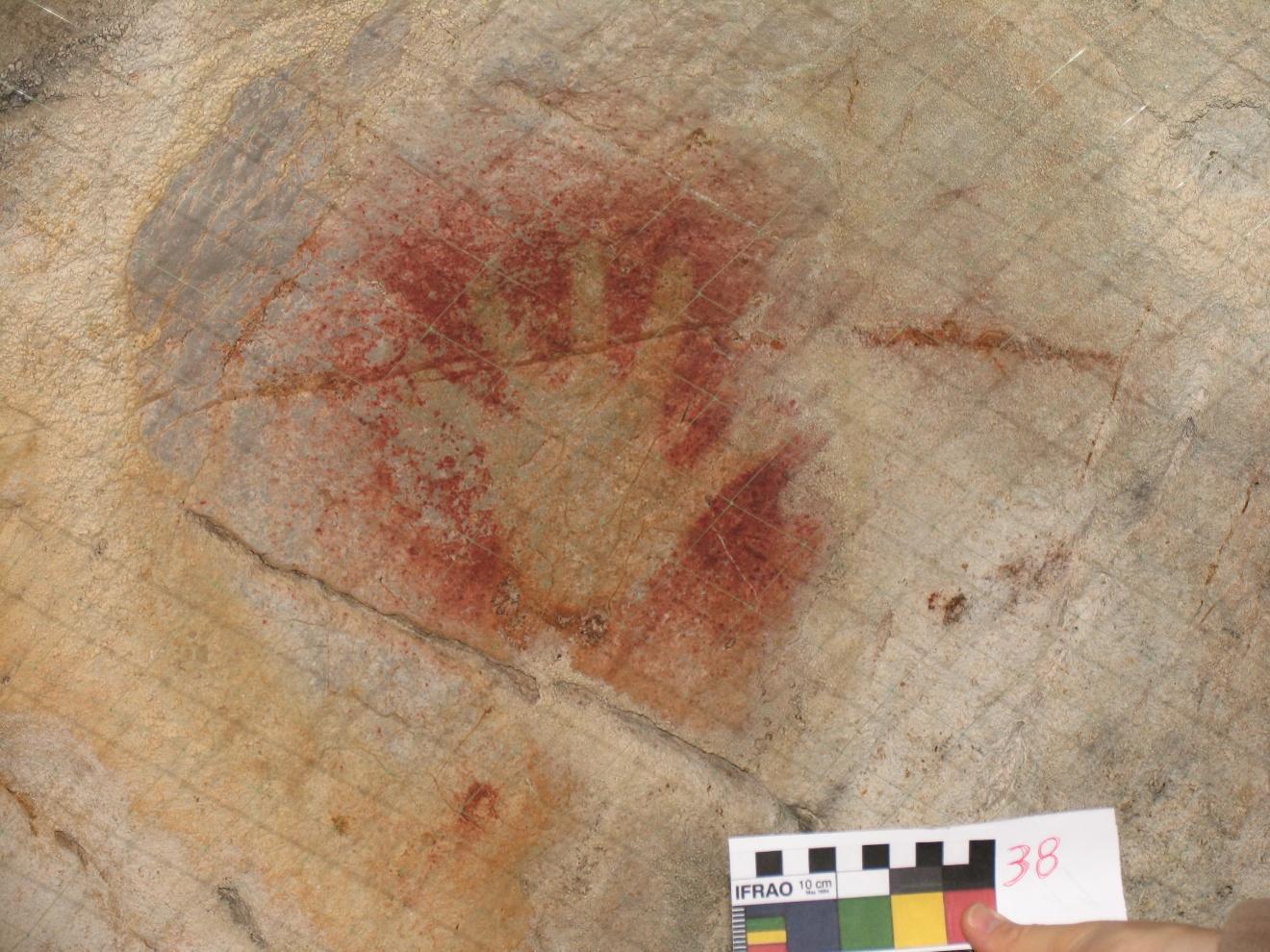}
		\caption{img\_38}
		\label{fig:img38}
	\end{subfigure}
	\caption{Four prehistoric hand images from El Castillo cave \parencite{snow:2006}.}
	\label{fig:intro_four_hands}
\end{figure*}

While DNNs offer significant promise, the analysis of prehistoric hand stencils remains affected by a range of archaeological and technical uncertainties. Pigment erosion, incomplete preservation, surface irregularities, uneven illumination, and poorly defined stencil boundaries can hinder both automated segmentation and manual contour delineation, with the latter being particularly sensitive to observer judgment and image noise. These challenges are illustrated in Figure~\ref{fig:intro_four_hands} using examples from El Castillo Cave \parencite{snow:2006}.

Representations based on photogrammetric 3D mo\-dels and 2D orthophotos can reduce viewpoint- and scale-related distortions, but they do not eliminate these inherent sources of uncertainty. Moreover, uncertainty may arise during 3D reconstruction and orthophoto generation: reconstruction errors can propagate to the final representation, while orthoprojection depends on the reconstructed surface geometry and the choice and orientation of the projection plane. Orthoprojection does not generally preserve intrinsic distances on curved or irregular cave surfaces, and the resulting distortions may vary across the same hand stencil according to local surface orientation. Although such distortions are small on nearly planar surfaces, they can introduce relevant geometric uncertainty into morphometric measurements on irregular ones. This issue is particularly relevant to sex-attribution methods, for which small variations in finger dimensions, hand proportions, or landmark locations can affect the extracted descriptors and subsequent classification. Finally, the absence of biological ground truth for Palaeolithic individuals represents an intrinsic source of uncertainty that cannot be resolved through geometric preprocessing alone.

Overall, these limitations, ranging from the absence of biological ground truth to surface degradation, underscore the need for approaches that explicitly account for uncertainty to support robust interpretations. Accordingly, methods that generate diversity through plausible geometric realizations of multiple source representations, identify patterns across multiple hypotheses, quantify uncertainty, and provide interpretable evidence for their decisions appear well suited to this task.

%Referencia a que estas limitaciones NO se resuelven con la introducción de  

%Within the deep learning framework, this work introduces a multi-layered, uncertainty-aware inference methodology. Uncertainty is explicitly modeled, propagated, diversified, and hierarchically aggregated across several analytical pipeline strata: the source level (via dual image processing), the annotation level (dual contour extraction), the morphological level (structured silhouette perturbations), the model level (architectural diversity and stochastic replication), and the decision level (ensemble fusion with silhouette support rates). By introducing controlled variability at each stage, the framework ensures that final predictions emerge from the convergence of multiple, structured, yet inherently imperfect observations.

This study proposes a multi-layered, uncertainty-aware inference methodology that models, propagates, and aggregates variation hierarchically across a deep learning pipeline. Building upon the classifier ensemble and spatial attribution-based interpretability method proposed in \parencite{mollineda:2025}, the present work substantially expands the previous framework in both methodological scope and uncertainty management. The extended framework systematically incorporates variability across all stages of the analysis: source-level variation through dual image processing, annotation-level variation through dual contour extraction, morphological variation through structured silhouette generation, uncertainty quantification at multiple stages, an additional interpretability tool ba\-sed on latent-space analysis and manifold mapping, and a triangulated validation scheme designed to identify convergent evidence across predictions, interpre\-ta\-bi\-li\-ty findings, and uncertainty measures. By propagating controlled variability throughout the pipeline, the proposed methodology goes beyond the previous ensemble-based approach, deriving final classifications from the convergence of multiple, highly structured yet individually imperfect observations.

Nine hand-stencil images are considered as case studies to demonstrate the practical applicability of the proposed methodology and its value as a tool to support archaeological interpretation. The cases are analyzed jointly through their sex predictions, associated confidence measures, latent-space organization, and spatial attribution patterns, providing complementary evidence for interpreting the model outputs. Importantly, the nine cases encompass a broad range of outcomes, from highly consistent predictions supported by multiple sources of evidence to cases characterized by substantial uncertainty or disagreement. This diversity enables a detailed examination of different analytical scenarios and illustrates how the methodology can support the interpretation of both well-supported and ambiguous cases. These nine images are not intended to provide a representative sample of the morphological variability of Upper Palaeolithic hand stencils; rather, their sole purpose is to demonstrate the practical value and analytical capabilities of the proposed methodology.

%five core stages: source-level (dual image processing), annotation-level (dual contour extraction), morphological-level (structured silhouette variations), model-level (architectural diversity and stochastic replication), and decision-level (classifier ensemble via silhouette support rates). 

The main contributions of this work are summarized below:%\\[-7mm]

\begin{itemize}
    \item A novel uncertainty-informed computational fra\-me\-work for sex attribution in prehistoric hand stencils, which explicitly models, propagates, and aggregates uncertainty across all stages of the analytical pipeline.
    %An extension of deep learning methodologies trained on modern medical data (pediatric hand X-rays) to the archaeological domain, enabling sex inference from prehistoric hand silhouettes under the absence of ground truth.
    \item A structured silhouette representation strategy that generates multiple plausible contour realizations through dual annotation, morphological perturbations, and multi-channel compositions, capturing annotation and morphological uncertainty.
    %A multi-level ensemble learning approach combining architectural diversity, stochastic model replication, and hierarchical aggregation, yielding robust predictions accompanied by interpretable confidence measures (e.g., silhouette support rate).
    \item A triangulated validation scheme integrating ensemble predictions, latent-space analysis (UMAP + k-NN), and aggregated interpretability maps (LayerCAM), providing convergent evidence and enabling the diagnosis of epistemic stability.
\end{itemize}

The remainder of this manuscript is organized as follows. Section~\ref{sec:related} reviews previous research on sex estimation from hand morphology and prehistoric hand stencils, with particular attention to the evolution from traditional morphometric approaches to contemporary machine learning and deep learning methodologies. Section~\ref{sec:materials-methods} introduces the contemporary and prehistoric datasets and the proposed inference framework, detailing the silhouette extraction and augmentation procedures, the ensemble-based classification strategy, and the post-hoc analysis techniques used for validation and interpretation. Section~\ref{sec:experiments} presents the experimental protocol adopted to benchmark the deep learning models on contemporary hand silhouettes and to deploy the framework on prehistoric hand stencils, including results obtained on both data domains. Section~\ref{sec:discussion} discusses the significance of the findings, the strengths and limitations of the proposed approach, and practical considerations for archaeological interpretation. Finally, Section~\ref{sec:conclusions} summarizes the main conclusions and outlines potential directions for future research.

\section{Related work}
\label{sec:related}

\begin{table*}[t]  % [hbtp]
	\centering
	\footnotesize
	\renewcommand{\tabularxcolumn}[1]{>{\centering\arraybackslash}m{#1}}
	%\newcolumntype{Y}{>{\centering\arraybackslash}p{1.6cm}}
	\begin{tabularx}{\textwidth}{XXXXXX} %|Y|Y|Y|Y|Y|Y|Y|}
	\hline\hline
	\textbf{Reference} & \textbf{Training data} & \textbf{Test data} & \textbf{Features extraction} & \textbf{Decision rule} & \textbf{Protocol} \\
	\hline\hline\noalign{\vspace{5pt}}
	\parencite{fernandez:2025b}  & Contemporary & Paleolithic & Automatically learned & DNN & Cross-domain\\[5pt]
	\hline
	%Fernandez-Navarro et al. 
	\parencite{fernandez:2025a}  & Contemporary & Simulated, Paleolithic & Automatically handcrafted & LDA &  Cross-domain\\[5pt]
	\hline
	%Fernandez-Navarro et al. 
	\parencite{mollineda:2025}  & Contemporary & Paleolithic & Automatically learned & DNN &  Cross-domain\\[5pt] % Held-out paleolithic test set %Paleolithic test samples
	\hline
	%Rabazo-Rodríguez et al. 
	\parencite{rabazo:2017} & Contemporary & Paleolithic & Manually handcrafted & LDA &  Cross-domain\\[5pt] % Held-out paleolithic test set %Paleolithic test samples
	\hline
	%Nelson et al. 
	\parencite{nelson:2017} & Contemporary & Contemporary & Automatically handcrafted & Fisher's LDA & Leave-one-out \\[5pt]
	\hline
	%Permana et al. 
	\parencite{permana:2015}. & Contemporary & Paleolithic & Manually handcrafted & Nearest class mean & Cross-domain\\[5pt] % Held-out paleolithic test set %Paleolithic test samples
	\hline
	%Galeta et al. 
	\parencite{galeta:2014} & Contemporary & Contemporary & Manually handcrafted & Fisher's LDA & Leave-one-out, Cross-domain \\[5pt] % Revisar: LOO es raro porque son conjuntos diferentes
	\hline
	%Wang et al. 
	\parencite{wang:2010} & Contemporary & Paleolithic & Automatically handcrafted & SVM & Cross-domain\\[5pt] % Held-out paleolithic test set %Paleolithic test samples
	
	\hline
	%Snow 
	\parencite{snow:2006} & Contemporary & Paleolithic & Manually handcrafted & Fisher's LDA & Cross-domain evaluation\\[5pt] % Held-out paleolithic test set %Paleolithic test samples
	
	\hline\hline
\end{tabularx}

\caption{Studies on sex attribution in prehistoric hand stencils, categorized by the type of hand data used (simulated stencils, Paleolithic stencils, or contemporary hand images). Abbreviations: LDA = Linear Discriminant Analysis; DNN = Deep Neural Network; SVM = Support Vector Machine.}
\label{tab:related_works}
\end{table*}

The analysis of hand stencils in rock art for sex estimation has evolved substantially, 
progressing from traditional anthropological measurements to more advanced computational 
approaches. In contrast to most reviews in this field, this section adopts a distinct perspective by focusing on four key dimensions: the nature of the hand data (Paleolithic stencils, simulated stencils, or contemporary), the feature extraction approach (manually handcrafted features, automatically extracted handcrafted features, or automatically learned features), the decision rule and the experimental protocol. Structuring the reviewed studies along these dimensions enables a clearer categorization and helps reveal common trends in hand stencil analysis. Table~\ref{tab:related_works} summarizes the selected studies, chosen for their experimental designs combining hand feature extraction and automatic sex classification.

One of the earliest and foundational studies was conducted by \parencite{snow:2006}, who introduced a two-stage analytical approach based on predictive discriminant analysis. Using a modern reference sample of 111 adult individuals of European descent, the first stage applied Fisher’s linear discriminant functions to overall hand length and  finger lengths (D2–D5) to separate adult males from individuals with smaller hands, a group that in archaeological contexts may include both adult females and subadults. Because absolute size alone cannot reliably distinguish women from young boys, the second stage applied the same discriminant framework to digit ratios (D2:D4 and D2:D5) to separate adult females from subadult males. The predictive equations, learned from the contemporary training set, were then applied to a small sample of six Paleolithic stencils, yielding one of the first demographic interpretations of cave artists.

%Expanding on this, \parencite{wang:2010} introduced a more automated machine learning pipeline using 175 contemporary hand images. Their methodology involved transforming images into the HSV color space and applying $K$-means segmentation to isolate hand contours. By identifying key points of interest—specifically fingertips and valleys—via curvature and angle signatures, they extracted 33 geometric features, including finger lengths, widths, palm dimensions, and digit ratios. To maintain consistency, features were normalized by the middle finger length (\colorbox{yellow}{excluding the thumb}), with a specific emphasis on the sexually dimorphic D2:D4 ratio. A Support Vector Machine (SVM) trained on this modern dataset was tested on prehistoric handprints from French caves. This study stands as a pioneering application of machine learning within an archaeological context.

Expanding on this, \parencite{wang:2010} introduced an automated machine learning pipeline using contemporary hand images. Their methodology involved transforming the images into the HSV color space and applying $K$-means clustering to isolate hand contours. By identifying key points of interest, specifically fingertips and valleys, they extracted geometric features including finger lengths, finger widths, palm dimensions, and the well-known D2:D4 ratio. To reduce the effect of scale differences across images, these measurements were normalized by the length of the middle finger. The middle finger was omitted after normalization due to its constant value, and the thumb was excluded due to unreliable measurements. Finally, a Support Vector Machine (SVM) classifier, trained on modern data represented as 33-dimensional feature vectors, was applied to prehistoric handprints from French caves. This study constitutes one of the earliest applications of machine learning to sex estimation in archaeological handprints.

The issue of population variability and methodological reliability was directly addressed by \parencite{galeta:2014} using an independent modern French data\-set of 100 right-hand prints from the University of Bordeaux. In a first analysis, linear discriminant functions based on hand and finger lengths (D2–D5) alongside the D2:D4 and D2:D5 ratios were determined directly on the French dataset using a leave-one-out cross-validation scheme. Second, to evaluate generalizability across populations, they applied pre-existing discriminant functions trained on a U.S. sample by \parencite{snow:2006} directly to the French dataset. The model trained on U.S. data exhibited poor generalization; differences in average hand size between the U.S. and French populations led to heavily biased classifications. Consequently, the authors concluded that modern morphometric references are unreliable for estimating the sex of prehistoric artists.

Concurrently, other researchers investigated the application of these methods across diverse geographical contexts. \parencite{permana:2015} employed the manually extracted D2:D4 digit ratio on both contemporary individuals—181 adults and juveniles from three villages—and hand stencils from Pettakere Cave in Indonesia. Their approach used a simple decision rule based on comparison with class means, applying a rule derived from contemporary data directly to archaeological samples. In a similar vein, \parencite{rabazo:2017} examined sexual dimorphism in hand stencils from El Castillo Cave (Spain). Their study relied on a contemporary dataset of 77 hand stencils, from which three key features were extracted: overall hand length, index finger length, and ring finger length. A discriminant analysis model was trained on this modern dataset and directly applied to classify 21 Paleolithic stencils.

A significant shift towards geometric morphometrics is evident in the work of \parencite{nelson:2017}. This study relied on a contemporary dataset of 132 hand stencils, capturing hand morphology through 19 two-dimensional landmarks. The authors applied Generalized Procrustes Analysis for shape alignment, followed by Principal Component Analysis for dimensionality reduction, and Fisher’s Linear Discriminant Analysis for classification. Their experimental protocol relied on cross-validation to estimate performance, due to the absence of an independent test set.

A comprehensive framework proposed by \parencite{fernandez:2025a} leverages a diverse, multi-source dataset to enhance the classification of ancient hand stencils. This dataset includes contemporary scans from 546 individuals (categorized by sex and age), 53 experimental stencils created in a limestone quarry using Paleolithic techniques (ochre and plant tubes), and 124 authentic Paleolithic stencils from eight Iberian caves. The methodology utilizes 32 anatomical landmarks to capture hand morphology. To isolate shape from external variables, Procrustes coordinates were computed to remove the effects of size, position, and orientation. Hand size was quantified via centroid size, while pairwise shape differences and variability were assessed using Procrustes distances and Principal Component Analysis, respectively. By applying discriminant analysis to these handcrafted features, the framework provides a robust and generalizable model for characterizing the authors of prehistoric rock art.

%A different strategy was later explored by \parencite{fernandez:2025b}, who used a supervised deep learning pipeline based on hand silhouettes. Contemporary hand images and experimental stencils produced under conditions similar to prehistoric techniques were first segmented with a ResU-Net. Several pretrained convolutional models were then trained and evaluated across ten independent rounds using a train-validation-test split restricted to the modern and experimental datasets. EfficientNetV2-S was identified as the most stable model and was subsequently applied to 45 archaeological stencils from Spanish caves. In contrast to morphometric analyses based on landmarks, linear measurements, and proportions, this strategy exploits the full hand shape through data-driven feature extraction. However, its applicability depends on stencil preservation quality and may be affected by the possible misclassification of children’s hands.

An alternative approach was explored by \parencite{fernandez:2025b}, who developed a supervised deep learning pipeline utilizing hand silhouettes. Their multi-source dataset comprised 247 contemporary hand scans, 11,076 images from the public 11K Hands database (representing 190 unique subjects with a female-to-male ratio of approximately 1.8:1), 61 experimental stencils produced on limestone by modern adults, and 45 Paleolithic stencils. Due to the complexity of the rock backgrounds, the experimental and Paleolithic stencils were manually segmented using Procreate\footnote{Procreate is a professional digital illustration app designed for iPad (https://procreate.com).}; conversely, silhouettes from the contemporary data were extracted automatically. To achieve this automation, a ResU-Net model \parencite{diakogiannis:2020} was initially trained on the 247 scanned hands and then applied to the 11K Hands dataset. To refine the model, 120 high-quality predictions from the 11K Hands collection were manually selected, added to the original training set, and used to retrain the architecture from scratch. After filtering for motion blur or shape-altering artifacts, 5,556 high-quality silhouettes were successfully extracted from the 11K Hands database. These were combined with the 247 scanned silhouettes and 61 experimental stencils to form the training and evaluation sets. The authors assessed several pretrained convolutional neural networks across ten independent train-validation-test splits. EfficientNetV2-S emerged as the most stable model and was subsequently deployed to predict the biological sex of the 45 archaeological stencils from Spanish caves. Beyond improving generalization, the combination of scanned and experimental data served to narrow the domain gap between modern and ancient samples; nonetheless, the model's outputs remain susceptible to variations in manual segmentation and the documented bias toward classifying subadult hands as female.

Existing literature has relied largely on handcrafted features, ranging from linear measurements and digit ratios to landmark-based geometric morphometrics. 
%Although these approaches have provided important insights, they require predefined anatomical features and may therefore limit the capacity of the model to capture more complex shape patterns. 
Recent studies have begun to move beyond this pa\-ra\-digm by using deep learning models trained on contemporary hand data and evaluated on archaeological samples. In particular, \parencite{mollineda:2025,fernandez:2025b} explore data-driven pi\-pe\-li\-nes where sex-discriminant features are automatically learned from hand silhouettes, moving away from hand-en\-gi\-neered linear measurements. This shift represents a key step toward more objective and flexible frameworks for the analysis of prehistoric hand stencils, while highlighting the need for robust cross-domain validation when models trained on modern data are applied to the Paleolithic field.

Despite these advancements, significant challenges persist. A fundamental limitation is the lack of definitive ground truth, as the biological sex of prehistoric stencils must be inferred from contemporary reference models rather than directly observed. Furthermore, while the classification pipeline utilizes hand silhouettes, those derived from Paleolithic art are inherently products of manual delineation. This process is susceptible to inter-observer variability and subjective bias, further compounded by factors such as pigment diffusion, rock texture, and varying states of preservation,  Consequently, subtle inaccuracies in the traced contours may propagate through the model, amplifying the uncertainty of the final predictions. An additional challenge stems from the cross-domain setting, where models trained on modern or experimental data are deployed on archaeological samples. %This domain shift is often exacerbated by limited demographic diversity in modern reference sets, which may lack the age range or morphological breadth found in prehistoric populations. 
Moreover, discrepancies in pose, acquisition conditions, and population-specific characteristics can hinder the generalization of learned features. Collectively, these challenges underscore the necessity for robust frameworks that account for segmentation variability, dataset representativeness, and domain adaptation, while moving beyond simple accuracy to provide reliable measures of predictive uncertainty.

%This study introduces a systematic deep learning framework for sex estimation in prehistoric hand art that explicitly accounts for uncertainty, where final predictions emerge from the statistical convergence of multiple diverse hypotheses rather than from a single model. To overcome the total absence of annotated ancient data, we leverage DNNs trained on contemporary hand silhouettes. To ensure high-confidence sex attribution, uncertainty is mitigated through a multi-stage pipeline of controlled variability, which initiates with data and annotation refinements (dual image processing and contour extraction) and extends to morphological robustness via structured silhouette perturbations. At the architectural stage, reliability is reinforced through multi-channel compositions, diverse model selections, and stochastic replication. Finally, individual outputs are integrated at the decision level via ensemble fusion and silhouette support rates. 

\section{Materials and methods}
\label{sec:materials-methods}

\subsection{Overview}

\begin{figure*}[t]
	\centering
	\includegraphics[width=1\linewidth]{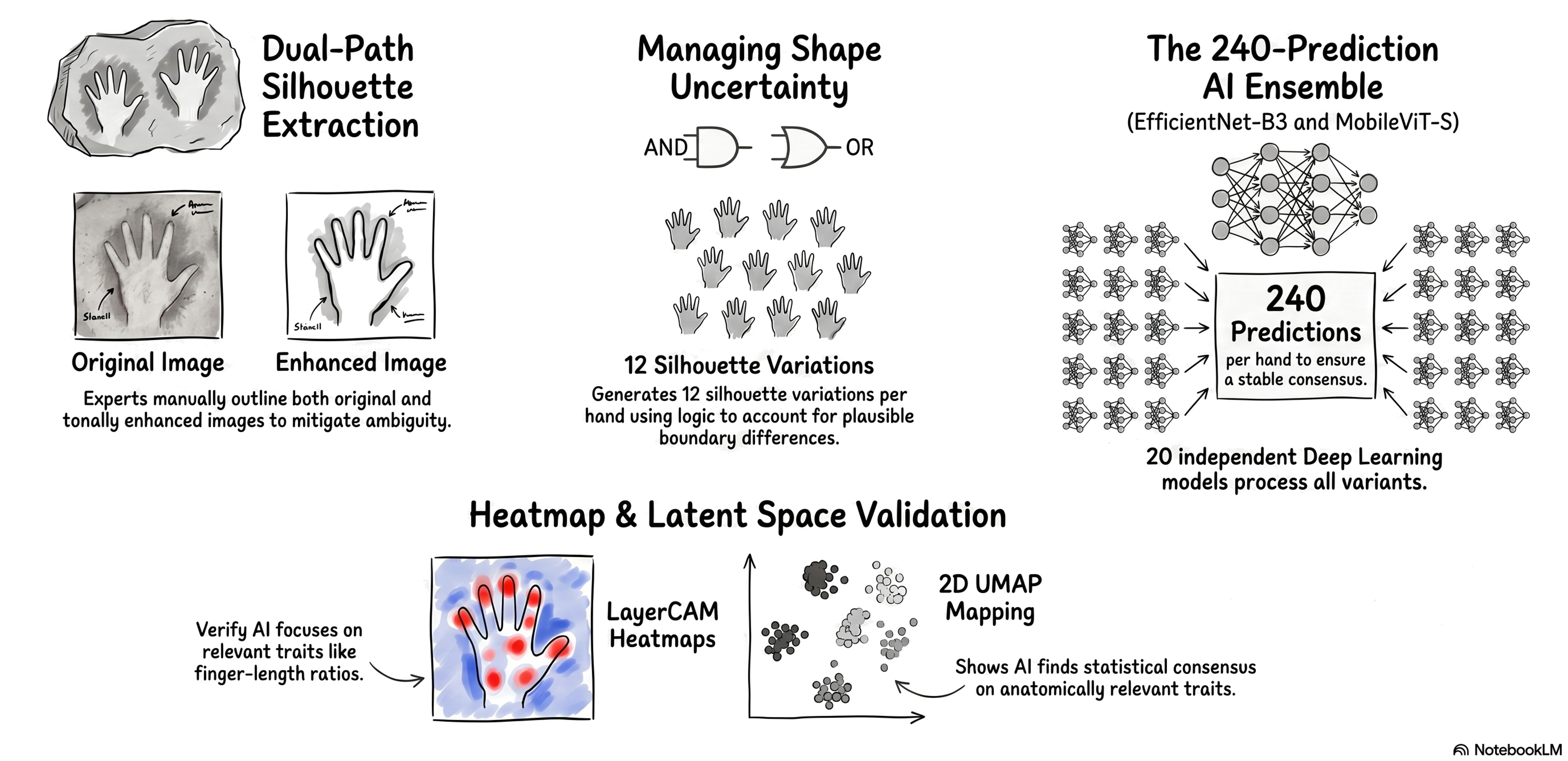}
	\caption{Uncertainty-aware inference system: quantifying sexual dimorphism in prehistoric cave hand stencils. Generated using Gemini Notebook from authors’ prompts and methodology.}
	\label{fig:methodology}
\end{figure*}

The proposed study aims to infer biological sex from prehistoric cave hand stencils through a sil\-hou\-ette-based computer vision pipeline,  %The approach integrates manual domain expertise, data augmentation, ensemble learning with two different DNN architectures, and post hoc analysis to ensure robustness, interpretability, and archaeological relevance. 
which is organized into three main phases: (i) silhouette extraction and augmentation, (ii) ensemble-based classification of silhouettes, and (iii) post-hoc analysis. Figure~\ref{fig:methodology} illustrates this computer vision pipeline.

\subsection{Datasets}

Experiments were conducted using two data sources: a large set of 14,036 annotated X-ray images from contemporary subjects, used to train DNNs, and a specialized collection of nine prehistoric hand-stencil images, used as case studies to illustrate the practical value of the proposed methodology. Samples from both sour\-ces were represented as binary hand silhouettes, enabling their integration within a unified experimental framework. This standardized representation supported the development of a DNN-based sex classifier trained on contemporary annotated silhouettes, which was then applied to the prehistoric data to generate sex-related hypotheses. %Although the training data derive from a pediatric cohort (including late adolescence, 16–19 years), the study assumes that DNN models capture morphological patterns of sexual dimorphism that remain informative across a broader age range.

\subsubsection{Contemporary data}

\begin{figure*}[t]
	\centering
	\includegraphics[width=\textwidth]{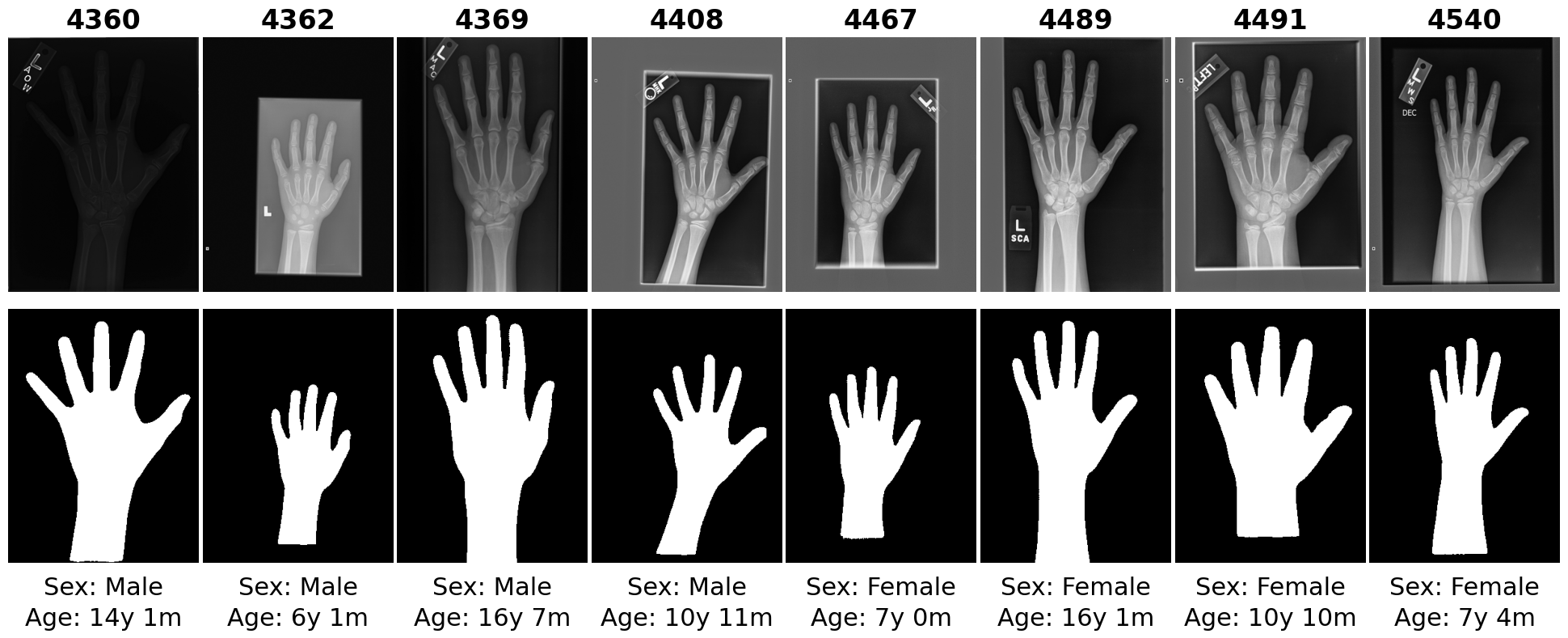}
	\caption{Representative samples of contemporary test X-ray images (top row) and their corresponding extracted silhouettes (bottom row). Each column displays a unique case identified by its 4-digit ID above, with associated sex and age metadata provided below.}\label{fig:test-cases}
\end{figure*}

\vspace{0.3cm}
This study relies on the RSNA Bone Age Challenge dataset \parencite{larson:2018,halabi:2019}, which comprises 14,036 X-ray images of the left hand from pediatric patients aged between 1 month and 19 years. The data were collected from two institutions: Lucile Packard Children’s Hospital at Stanford University (2,983 images) and Children’s Hospital Colorado (11,053 images) \parencite{larson:2018}. The dataset was randomly partitioned into 12,611 samples (approximately 90\%) for training and 1,425 samples (approximately 10\%) for validation. In addition, a separate test set of 200 hand radiographs (100 male and 100 female) was constructed from the picture archiving and communication systems at Stanford, ensuring no overlap with the training and validation studies. Each radiograph is annotated with skeletal (bone) age and patient sex based on the associated clinical radiology report. Silhouettes were obtained using the Segment Anything Model (SAM) \parencite{kirillov:2023} with a zero-prompt segmentation approach \parencite{mollineda:2025}. Table~\ref{tab:rsna-dataset} summarizes the sample counts and class distribution, while Fig.~\ref{fig:test-cases} presents representative test cases across both sexes and age groups.

\begin{table}[t]
	\centering
	\caption{Summary of RSNA Bone Age Challenge dataset.}
	\begin{tabular}{lccc}
		\hline\hline
		& females& males& total \\
		\hline
		training set   & 5,778 & 6,833 & 12,611 \\
		validation set & 652 & 773 & 1,425 \\
		test set       & 100 & 100 & 200\\
		\hline\hline         
	\end{tabular}
	\label{tab:rsna-dataset}
\end{table}

\begin{figure*}[t]
	\centering
	\includegraphics[width=0.32\textwidth]{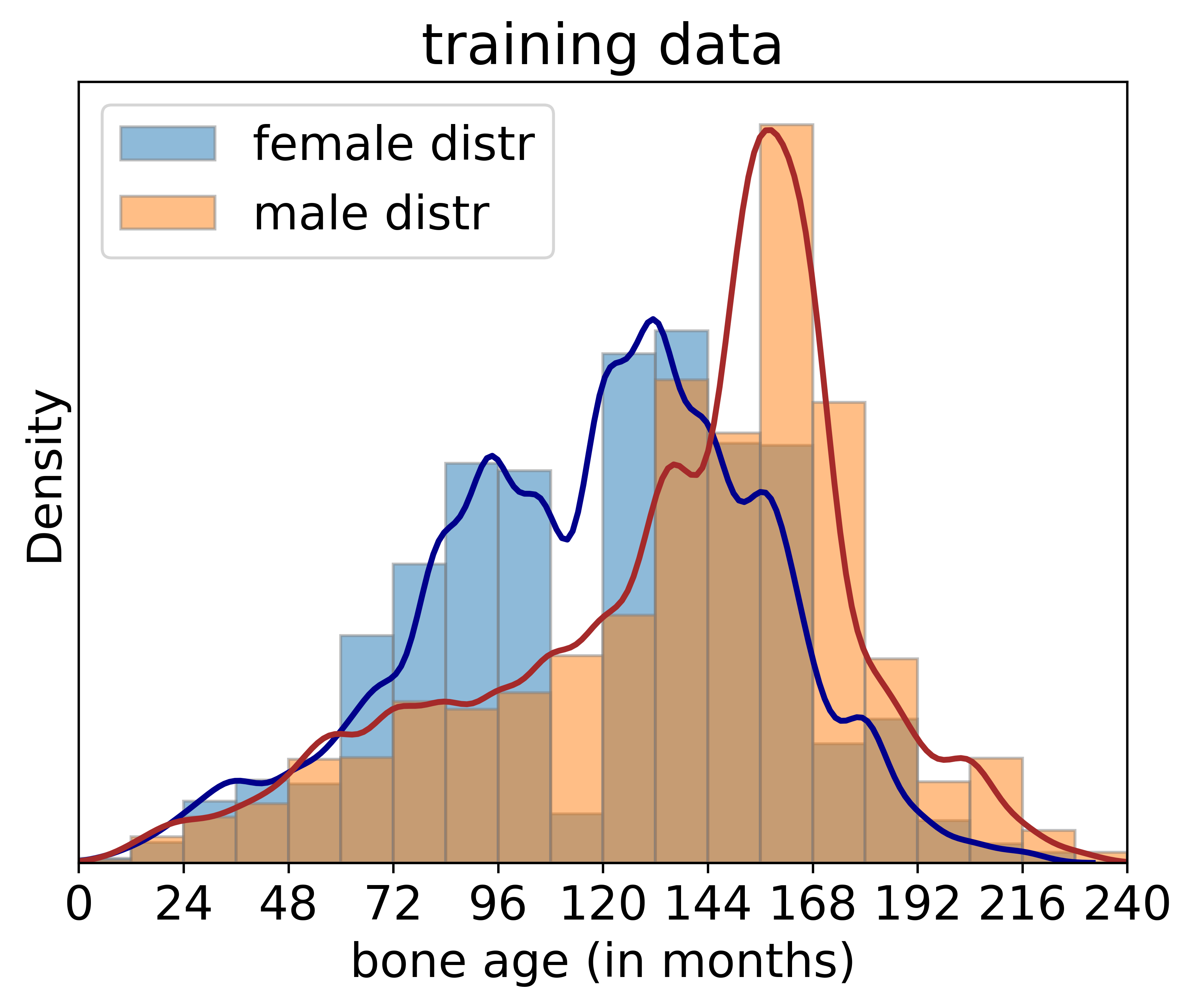}\hfill  %\hspace{-1mm}
	\includegraphics[width=0.32\textwidth]{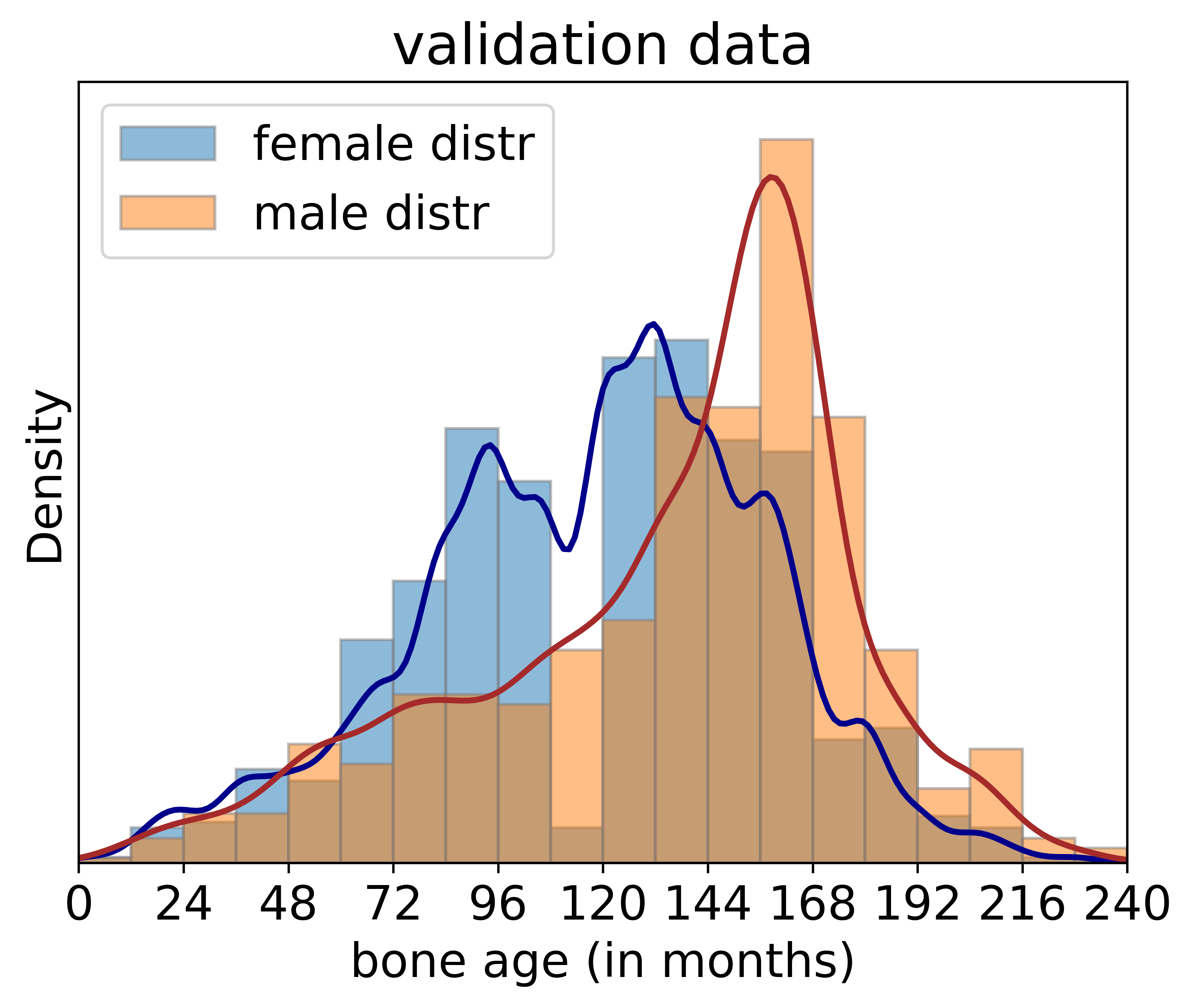}\hfill  %\hspace{-1mm}
	\includegraphics[width=0.32\textwidth]{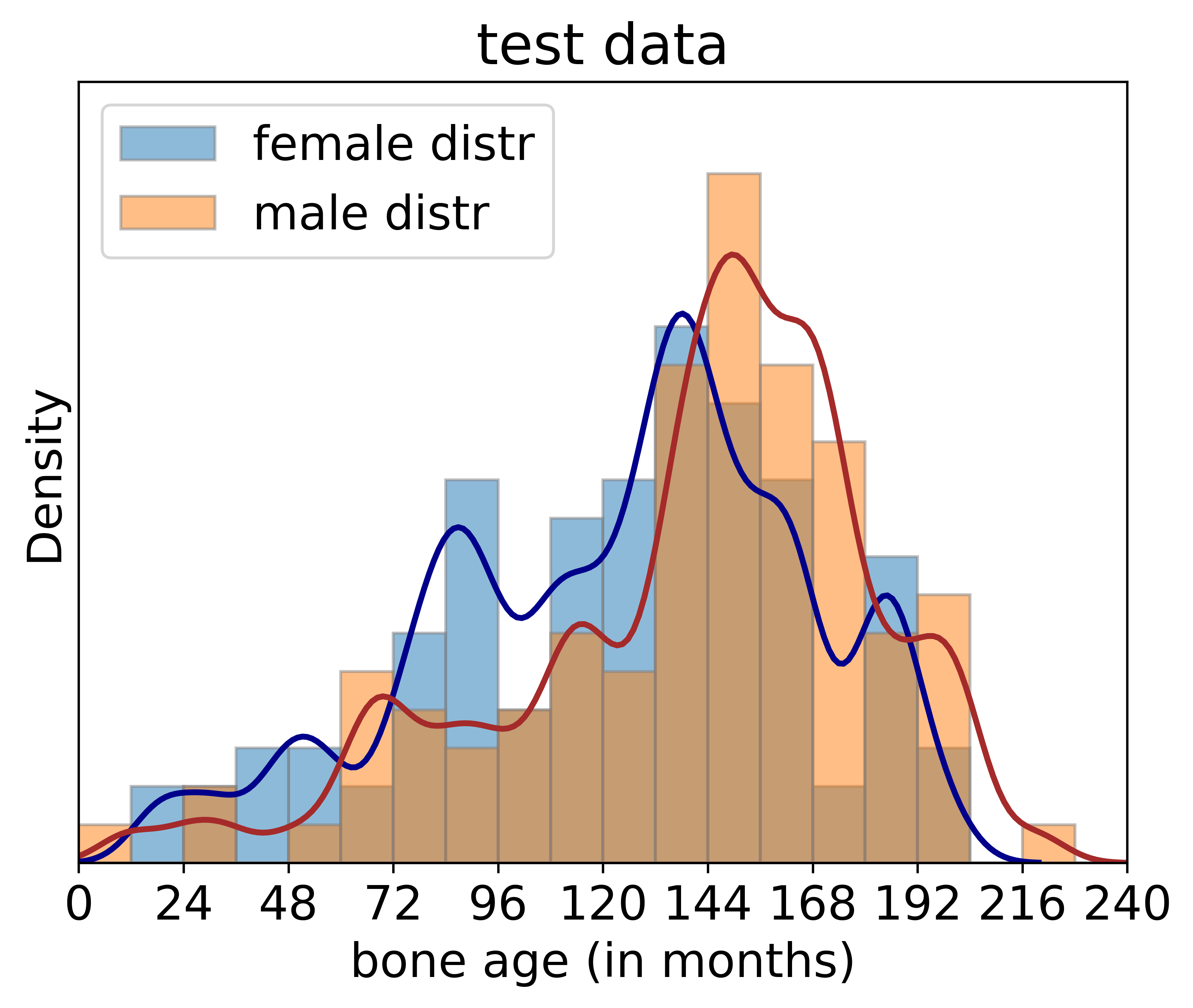}
	\caption{Bone age distribution by sex across the training, validation, and test subsets of the RSNA Bone Age Challenge dataset. Source: \parencite{mollineda:2025}. Reproduced with permission.}\label{fig:age-distr}
\end{figure*}

Figure~\ref{fig:age-distr} illustrates the class-conditional bone age distributions for the training, validation, and test sets. While the male and female classes exhibit unimodal and bimodal patterns, respectively, the highest concentration of samples lies between 10 and 16 years of age for both sexes. This age range corresponds to the attainment of skeletal and structural maturity of the hand, where sexually dimorphic traits become more stable and pronounced. The inclusion of a wide range of ages and morphological variability is expected to enhance the robustness and generalization capacity of the learned models.

\subsubsection{Prehistoric data}

\vspace{0.3cm}

\begin{figure*}[htbp]
	\centering
	% First Image
	\begin{subfigure}[b]{0.19\textwidth}
		\centering
		\includegraphics[width=\textwidth]{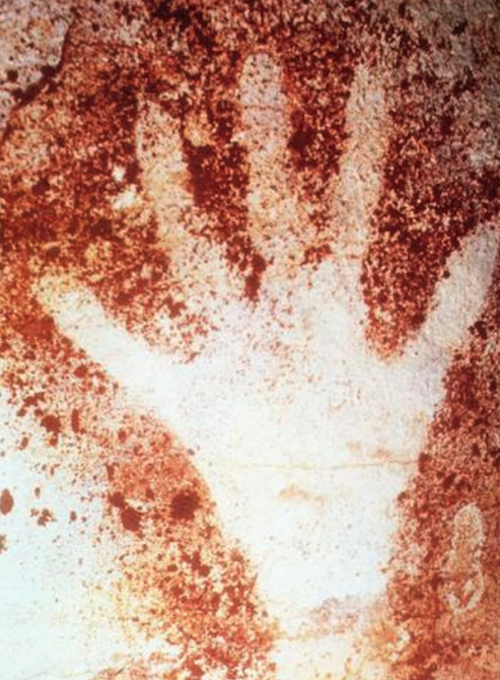}
		\caption{Cosquer}
		\label{fig:img_100}
	\end{subfigure}
	\hfill
	% Second Image
	\begin{subfigure}[b]{0.19\textwidth}
		\centering
		\includegraphics[width=\textwidth]{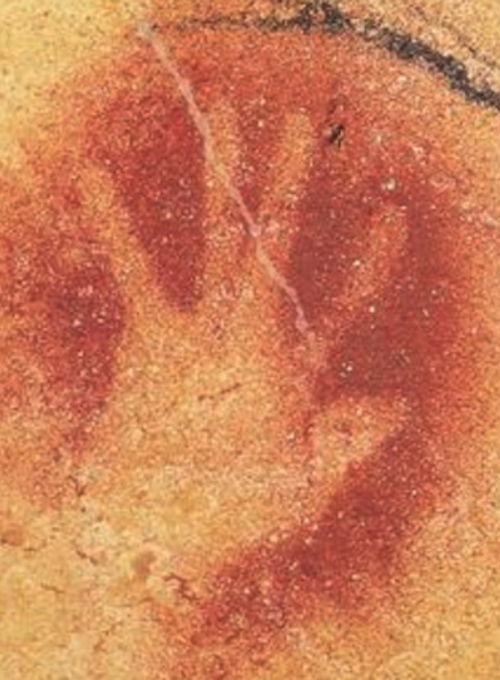}
		\caption{Chauvet}
		\label{fig:img_200}
	\end{subfigure}
	\hfill
	% Third Image
	\begin{subfigure}[b]{0.19\textwidth}
		\centering
		\includegraphics[width=\textwidth]{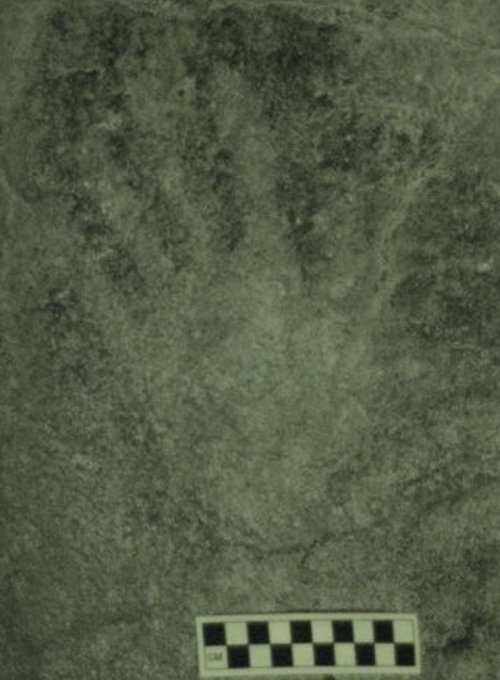}
		\caption{Unknown}
		\label{fig:img_201}
	\end{subfigure}
	\hfill
	% Fourth Image
	\begin{subfigure}[b]{0.19\textwidth}
		\centering
		\includegraphics[width=\textwidth]{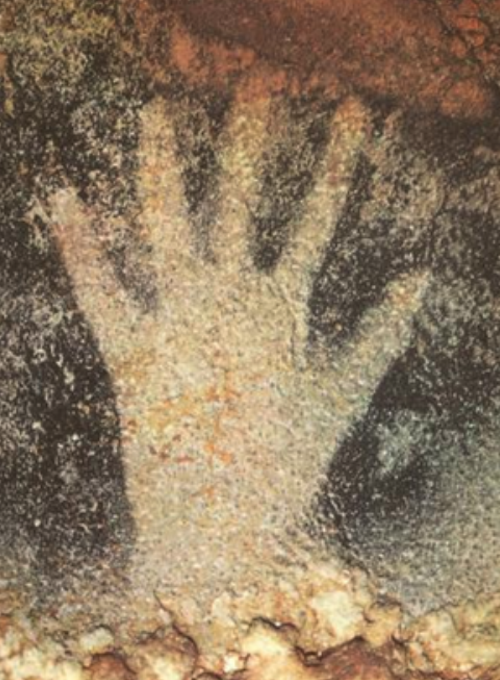}
		\caption{Pech Merle}
		\label{fig:img_202}
	\end{subfigure}
	% Fifth Image
	\begin{subfigure}[b]{0.19\textwidth}
		\centering
		\includegraphics[width=\textwidth]{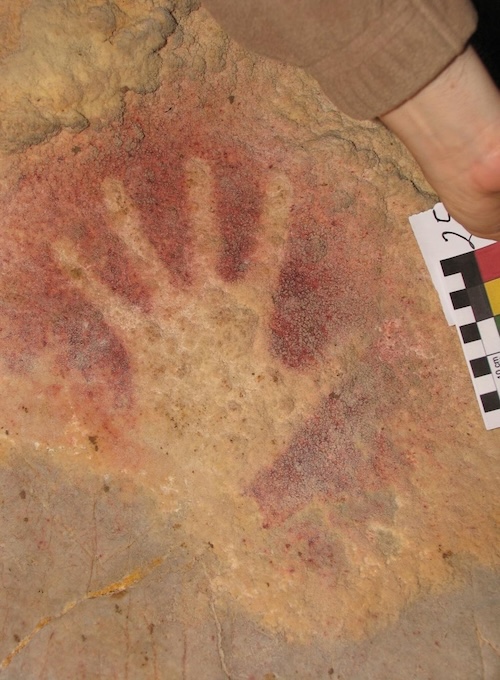}
		\caption{El Castillo}
		\label{fig:img_20325}
	\end{subfigure}     
	\caption{Five prehistoric hand images analyzed in the exploratory study \parencite{mollineda:2025}. Labels identify the source caves.}
	\label{fig:five_hands}
\end{figure*}

A total of nine prehistoric hand stencils from two distinct sources are used as independent case studies. The first subset comprises four samples from El Castillo cave (Spain), illustrated in Figure~\ref{fig:intro_four_hands}, which were selected for their superior contrast and well-defined edge morphology. These characteristics facilitate highly reliable manual contouring during the silhouette extraction process. The second subset consists of five samples previously analyzed by \parencite{mollineda:2025}, originating from at least four well-known sites (Cosquer, Chauvet, Pech Merle, and El Castillo). These images, shown in Figure~\ref{fig:five_hands}, were chosen because earlier studies proposed initial hypotheses regarding their sex attribution. Re-examining these cases under the substantially expanded uncertainty-aware framework introduced in this study is expected to provide more robust and reliable assessments, while offering a broader analysis of the evidence supporting each prediction. In addition, the use of nine hand stencils enables a detailed, case-by-case examination of the framework and illustrates how its different analytical components can support archaeological interpretation under conditions of substantial uncertainty. 

\subsection{Methodology}
\label{subsec:methodology}

\subsubsection{Overview}

\vspace{0.3cm}
The proposed methodology adopts a multi-layered, uncertainty-aware inference design to address key challenges in the analysis of Paleolithic hand stencils, particularly the lack of ground-truth information on the artists’ biological sex. Rather than attempting to eliminate uncertainty, the framework models, propagates, diversifies, and hierarchically aggregates uncertainty across the entire analytical pipeline. The uncertainty management strategies are summarized below across the three operational phases:

%\subsubsection{Uncertainty Management Framework}
% (i) silhouette extraction and augmentation, (ii) ensemble-based classification of silhouettes, and (iii) post-hoc analysis.

\begin{description}
	
	\item[Phase 1] Silhouette extraction and augmentation
	
	\begin{description}
	
	\item[Source-level uncertainty] At the archaeological and image-acquisition level, uncertainty arises from the absence of biological ground truth, surface degradation, pigment erosion, illumination variability, and historical interpretative disagreement. To mitigate photometric ambiguity, each stencil is processed along two parallel paths: the original image and a tonally enhanced version. This dual representation avoids committing to a single visual interpretation of degraded cave art.
	
	\item[Annotation-level uncertainty] Manual con\-tour extraction introduces boundary ambiguity and annotator bias. Instead of assuming a single definitive silhouette, the framework derives two independent contours from the original and enhanced images, yielding two plausible hand silhouettes. These are treated as alternative representations within a range of admissible boundaries rather than as exact geometries.
	
	\item [Morphological-level uncertainty] To explicitly model contour variability, structured shape perturbations are generated through binary operators: intersection (AND), union (OR), arithmetic averaging, and a morphological mid-set interpolation. In addition, three-channel compositional variants are built to account for representation dependency in convolutional architectures. In total, 12 silhouette representations are produced per stencil, expanding the morphological hypothesis space prior to classification.
	
	\end{description}
	
	\item [Phase 2] Ensemble-based classification of silhouettes
	
	\begin{description}
	
	\item[Model-level uncertainty] Epistemic and sto\-chastic uncertainties are addressed through architectural diversity and model replication. Two DNN architectures (Effi\-cient\-Net-B3 and MobileViT-S), each trained ac\-ross 10 independent runs with distinct random initializations, process all 12 silhouette variants, yielding a total of 120 probabilistic predictions per architecture (240 per stencil). This strategy intentionally amplifies variability prior to aggregation.
	
	\item[Hierarchical decision aggregation] For each architecture, predictions are aggregated using both sum-based and max-based fusion strategies. Consensus class labels are paired with a silhouette support rate, defined as the proportion of silhouette variants consistent with the predicted class. This metric serves as a morphological agreement index across contour realizations.
	
	\end{description}
	
	\item [Phase 3] Post-hoc analysis
	
	\begin{description}
	
	\item[Latent-space structural validation] To assess whether sex-specific structure is intrinsic to the latent space  learned by one the DNN architectures, latent feature vectors are projected into two-dimensional (2D) space using an unsupervised manifold learning technique. A non-parametric $k$-nearest neighbors ($k$-NN) classifier is then applied in the 2D embedding space. Agreement between ensemble predictions and 2D space classification indicates structural robustness under dimensionality reduction.
	
	\item[Interpretability-level aggregation] In\-ter\-pre\-ta\-bi\-li\-ty is assessed using attribution maps generated across silhouette variants and EfficientNet model instances. To suppress explanation noise and model-specific artifacts, 120 maps are aggregated via the geometric median, yielding a robust prototype explanation per stencil.
	
	\end{description}
	
\end{description}

%The proposed pipeline produces three complementary outputs for each prehistoric hand stencil image: robust sex predictions derived from multi-model score aggregation, latent-space visualizations and secondary $k$-NN classifications in a 2D embedding space, and model interpretability maps. By integrating ensemble-based scoring, manifold learning, and saliency mapping, the framework provides a comprehensive characterization of prehistoric samples while identifying the specific morphological cues driving the inferential outcomes.

In summary, the framework follows five key principles: (i) multiplicity before commitment, (ii) structured perturbation, (iii) architectural diversification, (iv) hierarchical aggregation, and (v) explicit quantification of internal agreement. The pipeline operates as a controlled uncertainty propagation and consensus system rather than as a single deterministic rule.

The next subsections provide a detailed description of the three phases outlined above.

\subsubsection{Phase 1: silhouette extraction and augmentation}

\begin{figure*}
    \centering
    \includegraphics[width=\textwidth]{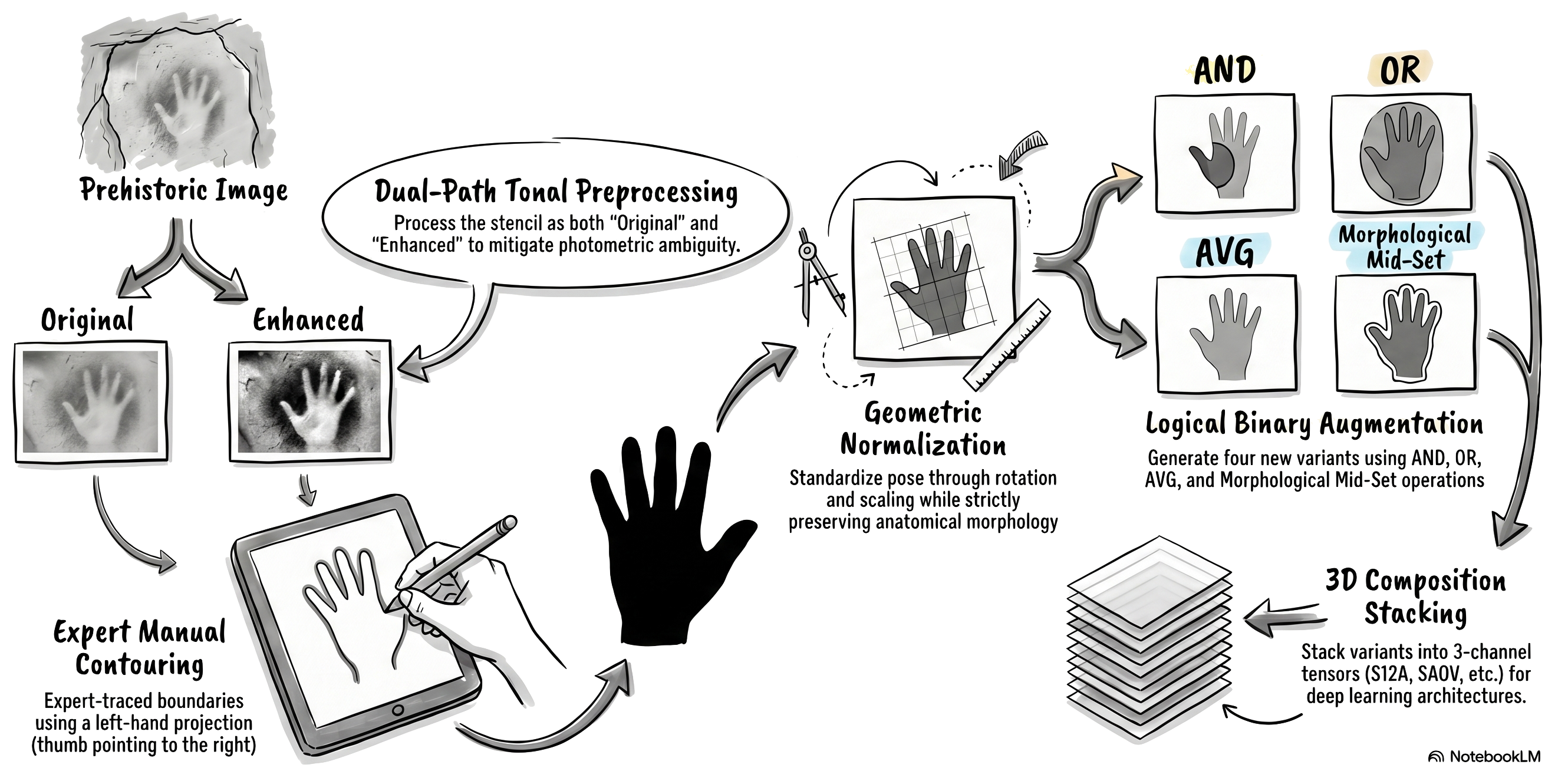}
    \caption{\textbf{Silhouette extraction and augmentation pipeline} (phase 1). From left to right: manual contouring is performed on both original and enhanced hand stencils, followed by the extraction of two primary silhouettes and their subsequent geometric normalization. The data augmentation phase comprises four binary operations (AND, OR, AVG, and morphological mid-set) between the extracted silhouettes. Finally, 3D composition operators combine the six resulting variants (two primary + four binary augmentations) into three-channel representations. Generated using Gemini Notebook from authors’ prompts and methodology.}
    \label{fig:silh_extract}
\end{figure*}

\vspace{0.3cm}
This phase comprises a multi-stage pipeline (see Fig.~\ref{fig:silh_extract}) encompassing two preprocessing operations: silhouette extraction and silhouette augmentation. Silhouette extraction transforms raw hand stencils into normalized, binary silhouettes, while silhouette augmentation manages manual tracing uncertainty by generating anatomically consistent, locally perturbed contour variants through a series of controlled spatial and pixel-wise transformations.

\begin{figure*}[t]
	\centering
	% First Image
	\begin{subfigure}[b]{0.25\textwidth}
		\centering
		\includegraphics[width=\textwidth]{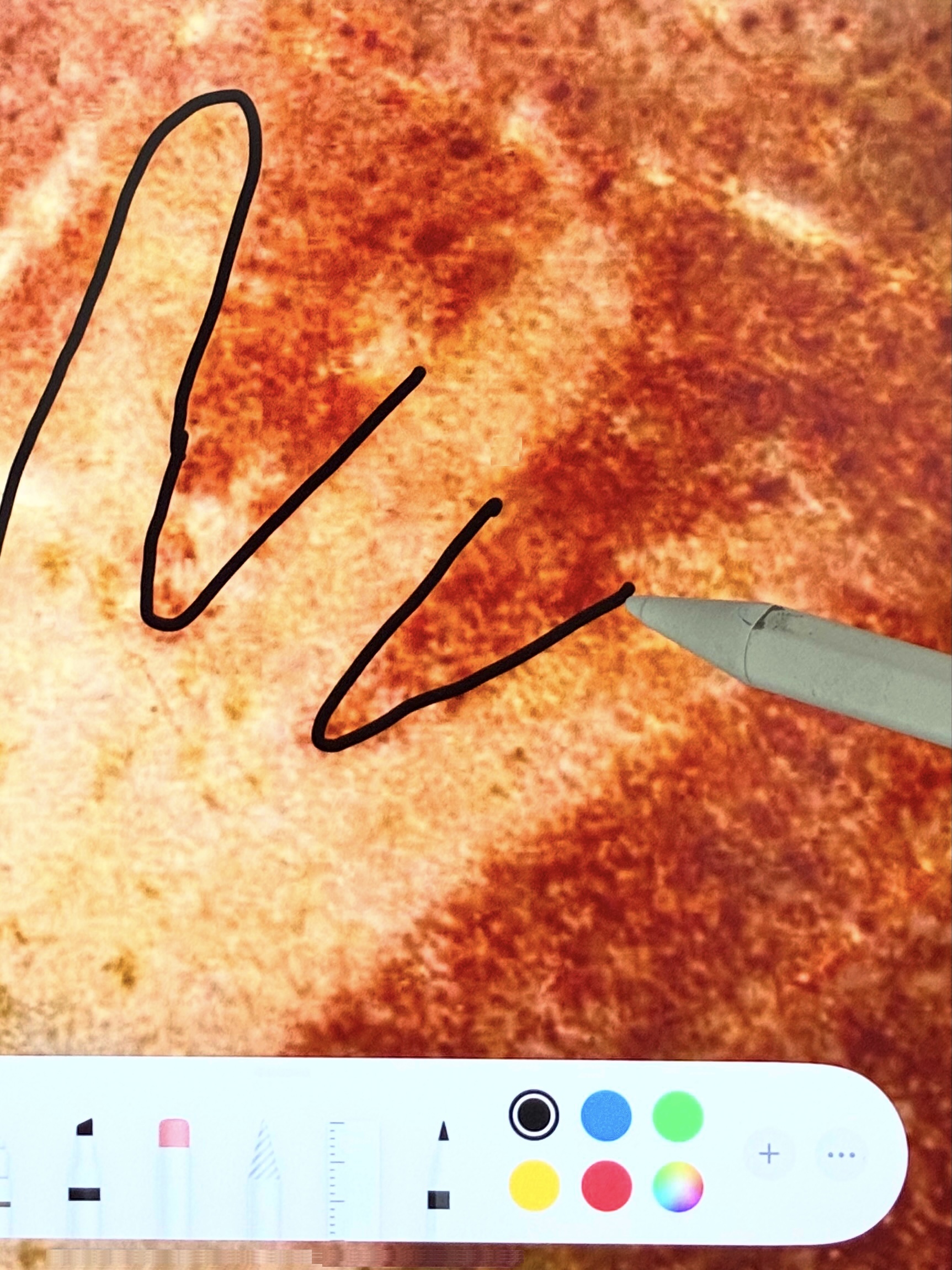}
		\caption{Manual contouring}
		\label{fig:contouring}
	\end{subfigure}
	\hspace{2mm} %\hfill
	% Second Image
	\begin{subfigure}[b]{0.25\textwidth}
		\centering
		\includegraphics[width=\textwidth]{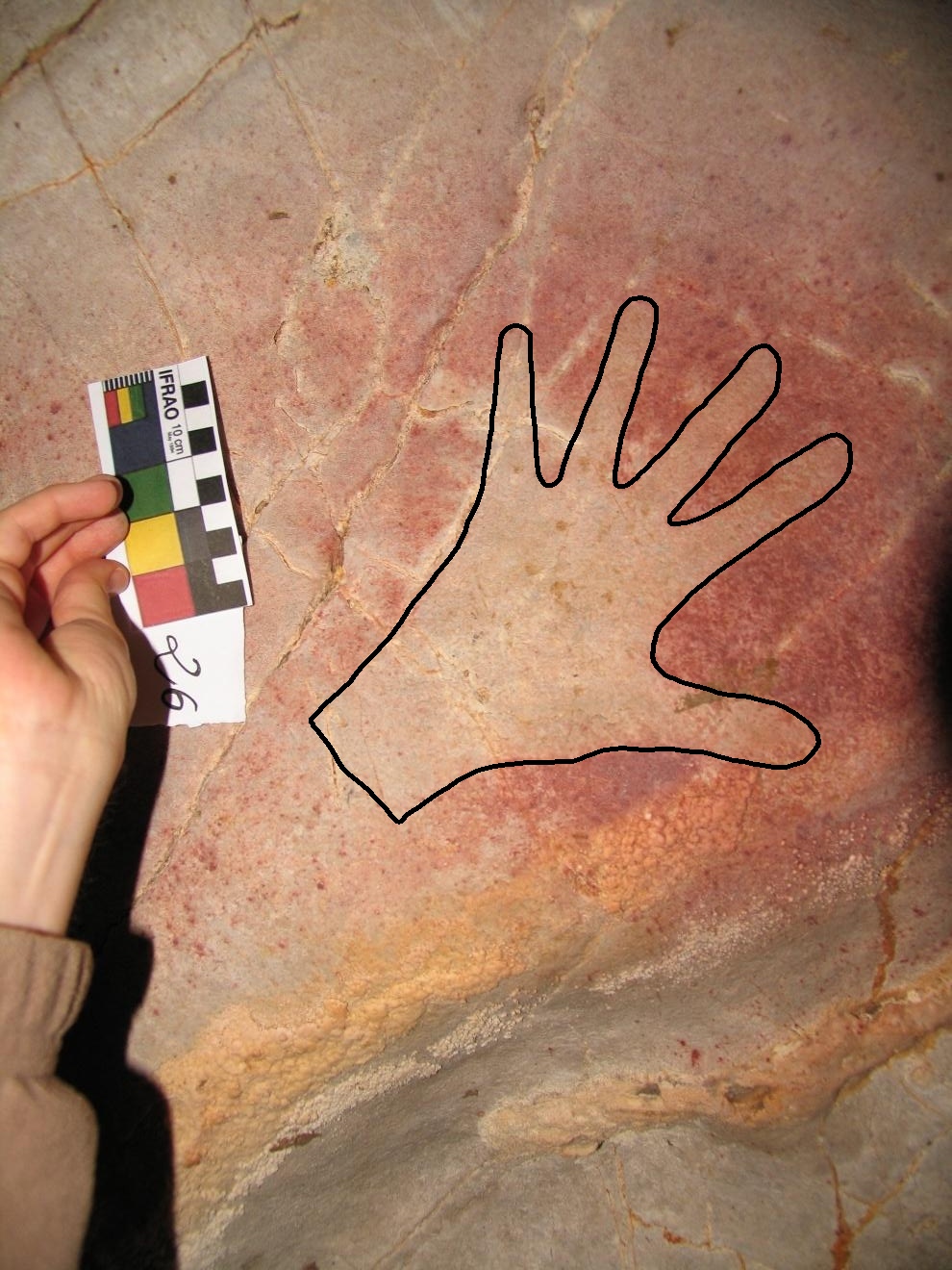}
		\caption{Original hand stencil}
		\label{fig:img26_s1}
	\end{subfigure}
	\hspace{2mm} %\hfill
	% Third Image
	\begin{subfigure}[b]{0.25\textwidth}
		\centering
		\includegraphics[width=\textwidth]{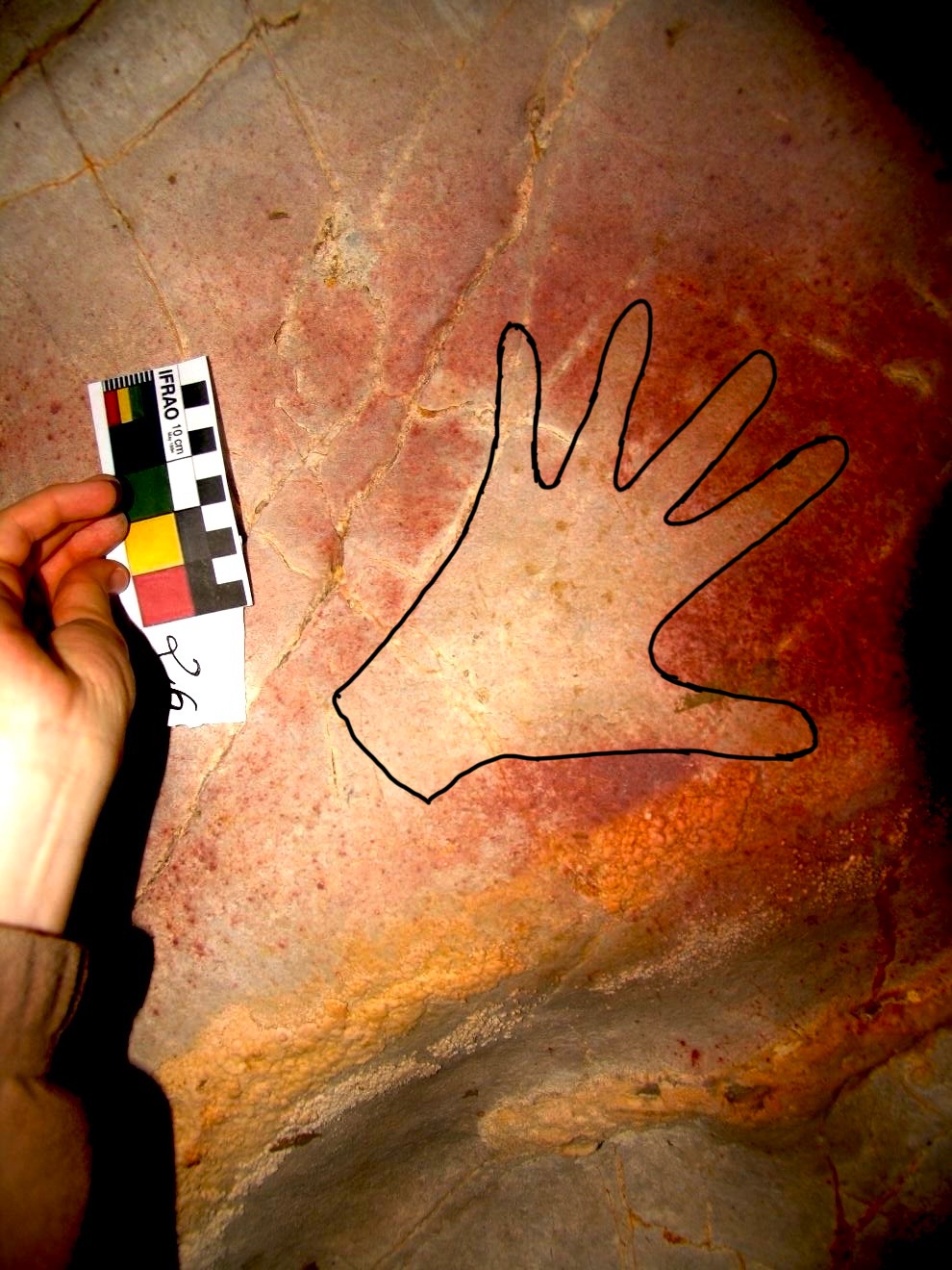}
		\caption{Enhanced hand stencil}
		\label{fig:img26_s2}
	\end{subfigure}
	\caption{Manual contouring applied to original versus enhanced cave hand stencil images.}
	\label{fig:manual_contour}
\end{figure*}

\textbf{Silhouette extraction}. The silhouette extraction process consists of several key steps:
 
\begin{itemize}
    \item \textbf{Adjustment of tonal image parameters (so\-urce-level uncertainty management)}. Cave art images typically present substantial challenges for automated segmentation due to uneven illumination, pigment degradation, surface irregularities, and background textures. The original image undergoes parametric adjustment through multiple enhancement areas: exposure, brightness, contrast, highlights, shadows, and sharpness. The methodology then branches into two separate paths, applying silhouette outlining and extraction to both the original and enhanced images to mitigate uncertainties arising from degraded image quality. 

    \item \textbf{Manual contouring (annotation-level uncertainty management)}. Hand silhouette extraction was achieved by manually contouring both original and enhanced cave hand images to isolate hand morphology. This dual-image approach (see Fig.~\ref{fig:manual_contour}) establishes a robust foundation for mitigating subjective annotation bias while providing complementary perspectives on ambiguous boundaries. By processing these two versions, the pipeline also compensates for the degraded quality and irregular pigmentation of ancient rock art.
    
    \item \textbf{Silhouette extraction}. The manual contours derived from the original and enhanced cave images feed into a hand silhouette extraction stage via contour masking. This process transforms outlined hand regions into clean binary hand masks (silhouettes) while suppressing background information. By decoupling hand morphology from photographic variability and surface irregularities, this approach enables standardized sha\-pe analysis in subsequent phases (see Fig.~\ref{fig:silh_extract}). This process comprises three primary steps: i) background suppression, ii) binary silhouette extraction, and iii) denoising and smoothing. Background suppression isolates the hand by zeroing out all background pixels. Binary silhouette extraction maps all hand pixels to white, yielding a binary image. Denoising and smoothing mitigate impulse noise via median filtering, while morphological operations eliminate artifacts, fill holes, and regularize contour boundaries.
  
    \item \textbf{Geometric normalization}. Manual axis alignment standardizes hand poses through rotation, scaling, and translation, ensuring consistency while strictly preserving anatomical morphology (see geometric in Fig.~\ref{fig:silh_extract}).

\end{itemize}

%\subsubsection{Silhouette augmentation: managing silhouette uncertainty}

\textbf{Silhouette augmentation}. Silhouette augmentation operations, defined from the two normalized hand silhouettes, explicitly account for both morphological and representational uncertainty.

Motivated by the uncertainty of manually contouring degraded, low-contrast images, silhouette augmentation acknowledges that contour placement is influenced by noise, surface deterioration, and subjective judgment; therefore, extracted silhouettes should be interpreted as plausible instances within a broader space of admissible shapes rather than precise geometries. From the two manually contoured silhouettes, augmentation generates locally perturbed variants via pixel-wise, morphological, and stacking operators, simulating other plausible contour variations without compromising global anatomy. 

From a learning perspective, each augmented silhouette represents a noisy observation (or a stochastic realization) of the underlying hand morphology. By evaluating all %stochastic 
variants with ensembles of silhouette-based sex classifiers and fusing their weak predictions, the approach promotes statistical consensus through aggregation: consistent morphological cues are reinforced across variants, whereas spurious or annotation-specific artifacts are averaged out. This strategy enables the final prediction to emerge from the convergence of multiple imperfect observations, yielding decisions that are less sensitive to local contour inaccuracies and more representative of the latent hand shape supported by the available visual evidence.

Let $S_1$ and $S_2$ be the two primary extracted binary silhouettes from the original and the enhanced hand images, respectively, such that $S_1,S_2\in\{0,1\}^{H\times W}$. Ten new augmented variants were generated from four binary and six 3D composition operators.

\textbf{Binary operators}. Silhouette augmentation is partially achieved using a set of binary operators detailed below:

\begin{itemize}
    \item \textbf{Pixel-wise AND} (logic): $S_{and}=S_1\cap S_2$. The AND operation yields an intersection silhouette, representing a high-confidence region where both silhouettes agree. It can be understood as a lower bound on the true hand shape, and referred to as ``consensus silhouette''.
    \item \textbf{Pixel-wise OR} (logic): $S_{or}=S_1\cup S_2$. The OR operation produces a union silhouette, forming a comprehensive shape envelope that includes all pixels from either silhouette. It could be an upper bound on the true hand shape, and referred to as ``composite silhouette''.
    \item \textbf{Average} (probabilistic): $S_{avg}=\frac{1}{2}(S_1+S_2)$. The average silhouette requires careful consideration since it transitions from binary to continuous-valued representation (soft silhouette). It can be interpreted as the probability that a pixel belongs to the true silhouette; thus, it can be called ``probabilistic silhouette''.
    \item \textbf{Morphological mid-set} (geometric): Let $S_{and}$ and $S_{or}$ be the inner and the outer silhouettes, such that $S_{and}\subseteq S_{or}$. The morphological mid-set is $S_{mid}=\{p\in S_{or}\;|\;d(p,S_{and})\leq d(p,{S}_{or}^c)\}$, where $d$ denotes Euclidean distance. Unlike statistical approaches, this geometric operation performs a distance-based shape interpolation, where each point in the output is closer (or equidistant) to the inner silhouette than to the outer boundary. The result is a geometrically intermediate shape that captures continuous transitions between contours.
\end{itemize}

\textbf{3D composition operators}. The silhouette-based sex classification DNNs were originally designed to process RGB images with three input channels. To adapt these architectures for grayscale or binary inputs, these 2D images were broadcast across three channels to construct a data structure equivalent to standard RGB data. This allowed for the full retraining of the DNNs on the new image domain. Consequently, the analysis of the original silhouettes ($S_1$, $S_2$), as well as their augmented versions ($S_{and}$, $S_{or}$, $S_{avg}$, $S_{mid}$), required replicating each silhouette across three channels prior to being presented to the models.

This architectural requirement motivated the exploration of alternative three-channel inputs, cons\-truc\-ted by stacking various silhouette variants—derived from the two primary hand silhouettes and their four augmentations—into composite 3D tensors. These stacked 3D compositions were then used as inputs to the pre-trained DNNs.

Within this pilot study, the following three-channel compositions were constructed and evaluated:
\begin{itemize}
    \item $S_{12A}=(S_1,S_2,S_{and})$
    \item $S_{12O}=(S_1,S_2,S_{or})$
    \item $S_{12V}=(S_1,S_2,S_{avg})$
    \item $S_{12M}=(S_1,S_2,S_{mid})$
    \item $S_{AOV}=(S_{and},S_{or},S_{avg})$
    \item $S_{AOM}=(S_{and},S_{or},S_{mid})$
\end{itemize}

In summary, for each cave image of a prehistoric handprint, a total of 12 silhouette representations were generated: $S_1$, $S_2$, $S_{and}$, $S_{or}$, $S_{avg}$, $S_{mid}$, $S_{12A}$, $S_{12O}$, $S_{12V}$, $S_{12M}$, $S_{AOV}$, and $S_{AOM}$.

\subsubsection{Phase 2: silhouette classification (sex prediction via score aggregation)}
\label{subsubsec:phase2}

\vspace{0.3cm}
In the absence of biological ground truth, sex prediction from cave hand stencils is formulated as a cross-domain classification task. Models are trained on a source domain with sufficient labeled samples (RSNA Bone Age Challenge dataset) and subsequently applied to a target domain lacking annotations (prehistoric hand stencils). Furthermore, the scarcity of high-quality samples in the target domain precludes the reliable use of domain adaptation techniques to mitigate distributional discrepancies between contemporary and prehistoric hand data.

Together, the lack of ground truth, the inability to reduce distributional shifts between source and target domains, and the ambiguity and subjectivity inherent in manually tracing prehistoric hand contours introduce significant uncertainty into individual model predictions.

To address this issue, this section proposes a classification strategy based on aggregating multiple weak (i.e., individually unreliable) predictions generated by ensembles of models applied to all silhouette variants. This approach integrates both representational diversity (across silhouette variants) and model diversity (across architectures and instances) into a consensus decision obtained through the fusion of numerous individual predictions. As a result, it enhances robustness to outliers and yields more stable classification outcomes.

Specifically, two ensembles of 10 model instances each are constructed from two complementary deep neural network architectures: EfficientNet-B3 \parencite{tan:2019} and MobileViT-S \parencite{mehta:2022}. The former provides a favorable performance–complexity trade-off for capturing fine-grained geometric details, whereas the latter employs a hybrid convolutional–transformer design to combine local feature extraction with long-range contextual modeling. This architectural diversity helps reduce systematic bias and promotes predictive variability within the ensemble.

Given a cave image, 10 EfficientNet-B3 and 10 MobileViT-S model instances process 12 distinct silhouette representations (2 manually contoured and 10 augmented variants), yielding 120 predictions per architecture (240 predictions per image in total). Each prediction is expressed as a probability distribution over two classes: \textit{Female} and \textit{Male}. For each architecture, predictions from the 10 independent model instances are aggregated using two  strategies:

\begin{itemize}
	\item \textbf{Sum aggregation}: Computes the cumulative posterior probability for each class across all ensemble members, favoring the class with the highest overall support.
	\item \textbf{Max aggregation}: Selects the maximum predicted probability for each class across all models, prioritizing the most confident individual prediction.
\end{itemize}

To ensure comparability, all model instances share a consistent standard training configuration: a batch size of 24, an initial learning rate of $0.001$, and 100 training epochs. The Adam optimizer is employed for its adaptive convergence properties. %To improve robustness, dropout (rate = 0.2) is applied to the fully connected layers, reducing sensitivity to specific silhouette variations. 
Training is guided by a cross-entropy loss with equal class weights, ensuring a balanced objective for binary classification. This configuration achieves stable convergence while maintaining strong generalization performance on the contemporary validation subset. To further enhance generalization and robustness to viewpoint and alignment variations, data augmentation is applied using moderate geometric transformations: rotation ($\pm 10^\circ$), width shifts of up to $10\%$, and height shifts of up to $10\%$. These transformations preserve the semantic integrity of the silhouettes while simulating realistic stochastic distortions.

%No explicit hyperparameter optimization has been conducted in this study. This decision is motivated by three considerations: (i) the primary contribution of this work lies in the introduction of an uncertainty-aware framework for classification and explainability, rather than in maximizing predictive performance; (ii) the methodological design intentionally prioritizes simplicity and accessibility to facilitate adoption by non-specialist communities; and (iii) achieving state-of-the-art performance on the related classification task falls outside the scope of this manuscript. Consequently, it is reasonable to expect that further improvements in both accuracy and robustness could be attained through a more systematic and rigorous hyperparameter optimization process.

Since comprehensive benchmarking for sex classification falls outside the scope of this work, explicit hyperparameter optimization was deliberately omitted in favor of developing an accessible, inference framework rather than maximizing predictive performance. This design choice helps ensure that the methodology remains user-friendly for non-specialists. Therefore, more thorough hyperparameter tuning represents a clear avenue for further improvements. % in both accuracy and robustness.

\subsubsection{Phase 3: post-hoc analysis and interpretation}
\label{subsubsec:phase3}

\vspace{0.3cm}
The final phase of the methodology focuses on post hoc interpretability and analysis, aiming to substantiate the model’s decision-making process. Specifically, this stage aims to provide insight into the internal mechanisms of the deep learning ensemble by examining both the geometric structure of feature representations in latent space and spatial attribution mapping. These post-hoc techniques foster model transparency and provide a rigorous means to explain and justify sex predictions through anatomically coherent evidence.

\textbf{Latent-Space Analysis and Manifold Mapping}. For each trained EfficientNet-B3 instance in the ensemble, the model automatically learns 1,536 high-level, sex-discriminant features. To make this high-dimensional data human-readable, these features are projected into a two-dimensional (2D) space using UMAP (Uniform Manifold Approximation and Projection) \parencite{mcinnes:2018}. This technique learns a nonlinear manifold in an unsupervised manner, meaning it reorganizes the data based on structural similarities without initially knowing the class labels. This dimensionality reduction serves two key purposes. First, it enables the visual assessment of the location of prehistoric hand stencils within the distribution of modern training samples. Second, it supports a secondary, non-parametric sex classification directly in the 2D space using the $k$-NN rule. 

\begin{figure}[t]
	\centering
	\includegraphics[width=\linewidth]{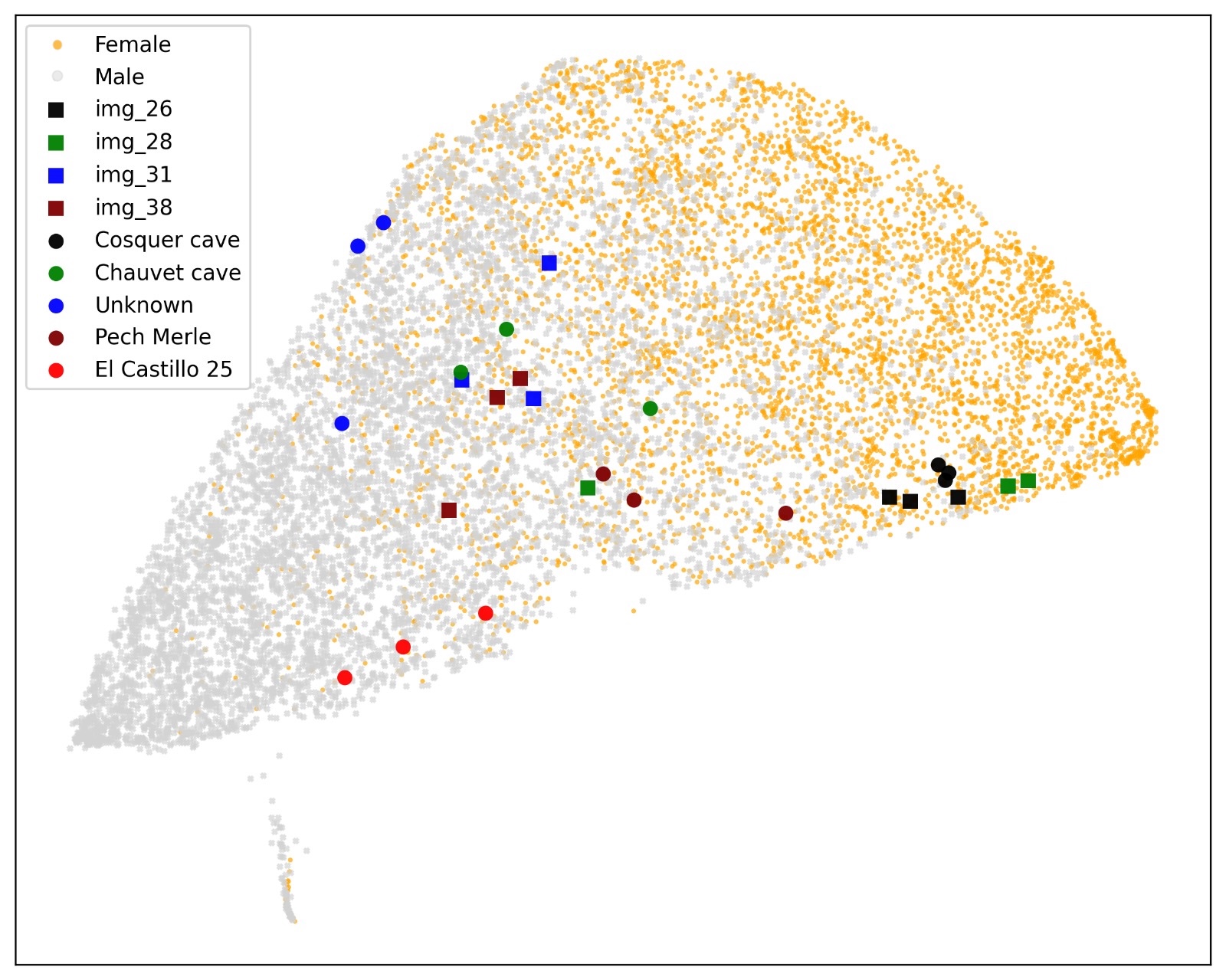}
	\caption{Interpretable 2D manifold generated by one of the ten EfficientNet-B3 models, using only the three silhouette variants $S_{and}$, $S_{or}$, $S_{mid}$ per stencil to reduce visual clutter.}
	\label{fig:umap}
\end{figure}

The 12 silhouette variants of the nine Paleolithic hand stencils are projected into each learned 2D space, yielding 108 projections per model (12 variants $\times$ 9 cave images). This provides a robust test set composed exclusively of samples designed to mitigate uncertainties associated with manual delineation. Figure~\ref{fig:umap} illustrates a representative 2D projection generated by one of the ten EfficientNet-B3 instances, using only three silhouette representations ($S_{and}$, $S_{or}$, and $S_{mid}$) per stencil to avoid visual clutter.

For each variant, a complementary classification is derived using a $k$-NN decision rule, utilizing the manifold-projected contemporary training silhouettes as reference data. These individual predictions are first combined via majority voting across variants to produce an instance-level consensus for each cave image. In a second aggregation step, majority voting across all ensemble instances (all manifolds) determines the final classification consensus. %Hyperparameter $k$ is empirically determined by assessing the average classification performance of the $k$-NN on contemporary validation data projected into the 2D manifolds.

\textbf{Model interpretability via spatial attribution}. Spatial attribution maps aim to evaluate whether model decisions rely on coherent and anatomically meaningful regions of the hand silhouette. The resulting heatmaps emphasize areas such as finger proportions and palm width that contribute to the predicted sex.

For each prehistoric hand image, attribution maps were generated via LayerCAM \parencite{jiang:2021} across the 12 silhouette representations and the 10 independent EfficientNet-B3 instances, yielding 120 maps per image. Therefore, these maps capture variability arising from contour uncertainty and model stochasticity. They were subsequently combined using the geometric median method \parencite{weiszfeld:1937}, building a single robust prototype attribution map per cave image, which emphasizes spatial patterns while suppressing outliers. Results are analyzed in Sect.~\ref{subsect:results-model-interpretability}.

\section{Experiments}
\label{sec:experiments}

This section is organized into two subsections to bridge the gap between controlled model validation and practical archaeological application. The first subsection evaluates the ensemble’s performance on contemporary hand silhouettes, establishing a performance baseline within the source domain where ground-truth labels are certain. The second subsection details the deployment of the inference framework on prehistoric hand stencils, the target domain of this study. By benchmarking the models on modern samples first, we provide a transparent and justified foundation for interpreting classification results and post-hoc analysis on the Paleolithic data.

\subsection{Contemporary data: model benchmarking on modern silhouettes}

\subsubsection{Experimental design}

\vspace{0.3cm}
A contemporary dataset of hand silhouettes with sex annotations was derived from the RSNA Bone Age Challenge, which contains 14,036 left-hand X-ray images from pediatric patients aged between 1 month and 19 years. Silhouettes were extracted using the Segment Anything Model (SAM) \parencite{kirillov:2023} through the zero-prompt segmentation strategy proposed in \parencite{mollineda:2025}. This approach achieved a silhouette detection sensitivity exceeding 99\% on the training and validation subsets, and 98\% on the test subset. 
%Table~\ref{tab:positive-silhouettes} summarizes the number and rates of silhouettes correctly detected by SAM (detection recall) in each subset of X-ray images, broken down by sex. 
Accordingly, the numbers of modern hand silhouettes used to train, validate, and test the DNN models are 12,509, 1,416, and 196, respectively. The broad age range is expected to facilitate the learning of robust sexual dimorphism features across all stages of skeletal development. Importantly, the highest concentration of samples occurs between 10 and 16 years for both sexes, a stage at which bone maturity is largely attained, thereby providing highly informative patterns for sex-related morphological differentiation. All images were resized to a $400\times 300$ resolution and normalized to the $[0, 1]$ range. 

%Both DNN architectures (EfficientNet-B3 and MobileViT-S) were implemented using the PyTorch library \parencite{paszke:2019} with weights pretrained on the ImageNet dataset \parencite{deng:2009}. 
%The EfficientNet-B3 and MobileViT-S architectures were implemented using the Torchvision and timm libraries, respectively, within the PyTorch framework (Paszke et al., 2019). Both models utilized weights pretrained on the ImageNet dataset (Deng et al., 2009). 
The EfficientNet-B3 and MobileViT-S architectures were implemented in PyTorch \parencite{paszke:2019} using the Torchvision and timm libraries, respectively, and initialized with ImageNet-pretrained weights \parencite{deng:2009}. Each model was adapted with a single fully connected output layer to predict the class-conditional probability distribution for each input sample. While initialized with pretrained weights to provide a meaningful starting point, all models were fully retrained on the hand silhouette images. Hyperparameter values remained constant across all classification tasks, as specified in Sect.~\ref{subsubsec:phase2}.

For each architecture, ten independent model instances were trained, sharing identical designs, initial weights, and hyperparameters. For each model instance, the training subset is used to optimize the DNN parameters over 100 epochs, the validation subset to select the optimal model checkpoint (i.e., the epoch yielding the highest validation accuracy), and the test subset to conduct a final independent evaluation of the selected model. 

The primary source of diversity across the ten individual model was the stochasticity in optimization introduced by the random ordering of training samples. Given the non-convex nature of deep learning loss landscapes, varying the sample sequence dictates the optimization trajectory through the high-dimensional weight space. This path-dependency causes models with identical configurations to converge toward different local minima and develop unique decision boundaries, thereby fostering the predictive diversity essential for a robust ensemble.

The resulting ensembles (one per architecture) aggregated the predictions of their ten constituent models using both sum and max rules. Classification performance was evaluated based on the overall accuracy (success rate).

Training and inference were conducted on an on-premise server equipped with an AMD Ryzen™ 9 7950X CPU, 64 GB of DDR5 RAM, and an NVIDIA GeForce RTX 4090 GPU (24 GB). The software environment included Ubuntu 22.04.4 LTS, Python 3.10.16, and PyTorch 2.5.1. %With an average training time of 42 s per epoch and 70 min per model, the total execution time was approximately 70 h.

\subsubsection{Sex prediction results on contemporary data}

\vspace{0.3cm}

\begin{table*}[t]
	\footnotesize
	\centering
	\caption{Performance of EfficientNet-B3 and MobileViT-S on contemporary data, featuring age-stratified results for the test set and overall results for training and validation sets. Results are reported as mean accuracy $\pm$ Standard Error of the Mean (SEM) for individual model instances, alongside ensemble accuracy achieved via sum and max aggregation rules. Square brackets indicate inclusion of the boundary values, whereas parentheses indicate their exclusion.}
	\vspace{1mm}
	\label{tab:model-benchmarking}
	\begin{tabular}{l@{\hskip 5mm}l@{\hskip 5mm}c@{\hskip 5mm}c@{\hskip 5mm}c@{\hskip 5mm}ccc}
		\toprule
		\multirow{2}{*}{Architecture} & \multirow{2}{*}{Model} & Training & Validation & \multicolumn{4}{c}{Test data} \\
		& & data & data & {All ages} & {[0--6) y} & {[6--12) y} & {[12--19] y} \\
		\midrule
        \multirow{3}{*}{EfficientNet-B3} & Mean$\pm$SEM & 
		$82.6\pm0.7$ & $75.8\pm0.3$ & $81.7\pm0.4$ & $68.1\pm1.9$ & $81.5\pm1.0$ & $85.3\pm0.6$ \\[0.5mm]
		& Sum aggregation & 
		$85.4$ & $77.7$ & $\mathbf{85.7}$ & $\mathbf{76.2}$ & $\mathbf{85.4}$ & $\mathbf{88.4}$ \\[0.5mm]
		& Max aggregation & 
		$\mathbf{85.9}$ & $\mathbf{78.1}$ & $84.7$ & $\mathbf{76.2}$ & $84.3$ & $87.2$ \\[2mm]
		%\hline
		\multirow{3}{*}{MobileViT-S} & Mean$\pm$SEM & 
		$80.3\pm1.4$ & $74.8\pm0.7$ & $80.6\pm1.7$ & $66.7\pm4.1$ & $79.4\pm1.9$ & $85.2\pm1.6$ \\[0.5mm]
		& Sum aggregation & 
		$84.2$ & $77.0$ & $82.7$ & $66.7$ & $82.0$ & $87.2$ \\[0.5mm]
		& Max aggregation & 
		$84.6$ & $77.8$ & $83.2$ & $61.9$ & $83.1$ & $\mathbf{88.4}$ \\[1mm]
		\bottomrule
	\end{tabular}
\end{table*}

The benchmarking of the models on modern silhouettes acts as a critical baseline to validate the ensemble's predictive performance within a source domain where ground-truth sex labels are certain. The results in Table~\ref{tab:model-benchmarking} demonstrate that both the Efficient\-Net-B3 and MobileViT-S architectures, and particularly their respective ensembles, exhibit a high capacity for sex prediction, achieving substantial accuracy across the contemporary test set. This performance is particularly pronounced in the [6–12) and [12–19] age groups, where the EfficientNet-B3 ensemble using sum aggregation achieves accuracies of up to $85.4\%$ and $88.4\%$, respectively.

A clear trend is observed whereby classification accuracy improves with increasing age of the subjects. This pattern appears to reflect two main factors. First, the dataset exhibits a higher sample density between 10 and 16 years, leading to greater representation in the older strata ([6--12) and [12--19]), which likely supports more robust learning of sex-discriminative features than in the youngest cohort [0--6). Second, morphological manifestations of sexual dimorphism become progressively more pronounced with age and are further consolidated during adolescence, yielding more distinct and informative skeletal patterns for model discrimination. This finding provides evidence for the models’ ability to reliably classify adult hands.
%As a particular case, the results obtained by aggregating the outputs of EfficientNet-B3 instances using the max rule are noteworthy, as they remain consistently strong and stable ($>85.0$) across all age strata.

Results also highlight the effectiveness of the ensemble approach, as these configurations consistently outperform the mean performance of individual model instances. This improvement is particularly pronounced for EfficientNet-B3 with both aggregation rules in the [0–6) age stratum, where the ensemble accuracy ($76.2\%$) substantially exceeds the average accuracy of individual instances ($68.1\% \pm 1.9\%$).

Across all datasets and test age strata, EfficientNet-B3 achieves higher average performance than Mobile\-ViT-S for individual models, and this advantage extends to most ensemble configurations under both aggregation rules. This behavior may be attributed to the greater architectural capacity of Effi\-cient\-Net-B3 (both in depth and parameterization) which enables the capture of more subtle morphological patterns. In addition, MobileViT architectures may be more sensitive to specific hyperparameter settings than the highly optimized and robust EfficientNet framework, particularly given that no hyperparameter tuning was conducted in this study. Nevertheless, these observations should be interpreted with caution and do not support broad claims regarding the general superiority of one architecture over another, as the analysis is restricted to specific model variants applied to a single, specialized task.

\begin{figure*}[t]
	\centering
	\includegraphics[width=0.75\textwidth]{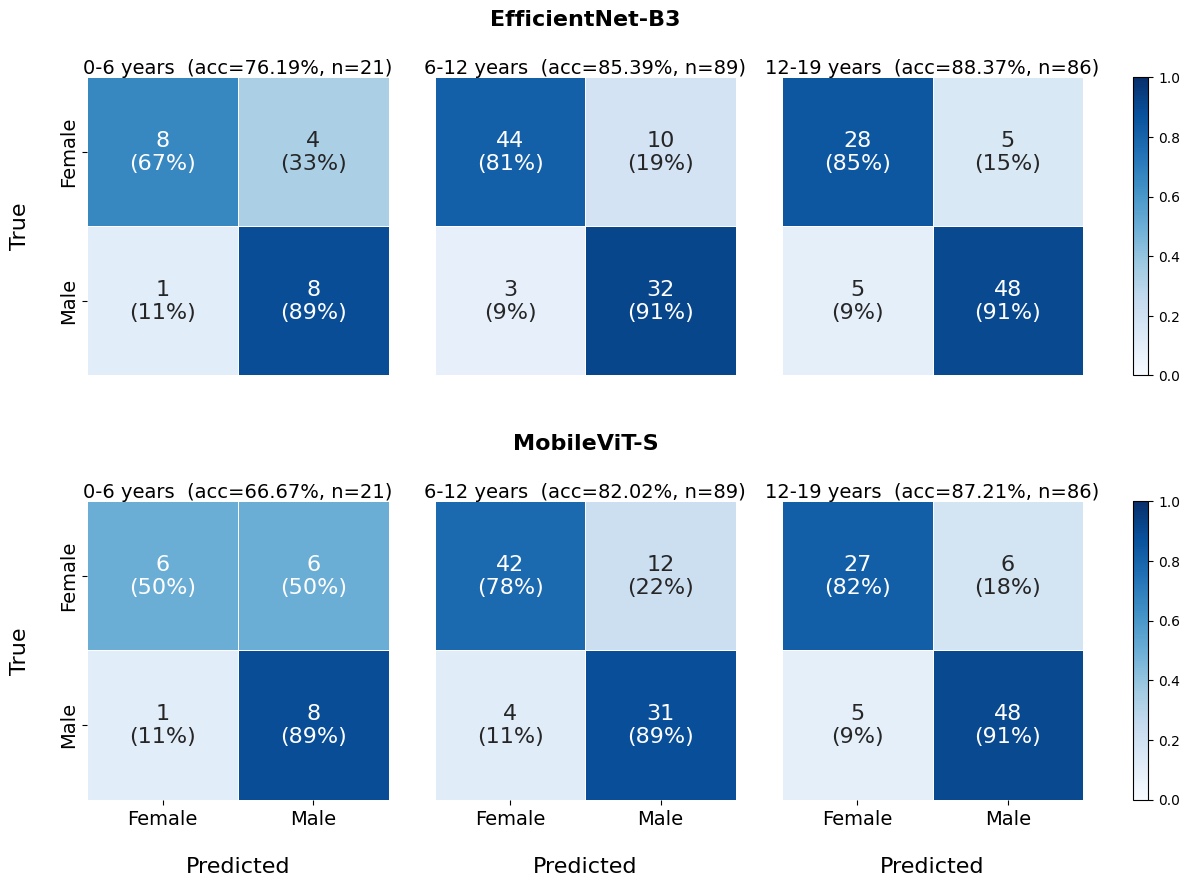}
	\caption{Confusion matrices summarizing class-wise correct and incorrect predictions across age-stratified test data for both ensembles (architectures) under the sum-aggregation rule.}
	\label{fig:confusion-matrices}
\end{figure*}

Figure~\ref{fig:confusion-matrices} presents the confusion matrices, detailing class-wise correct and incorrect predictions across age-stratified test data for both ensembles (architectures) under the sum-aggregation rule. They show a clear and coherent pattern across age groups and architectures, with classification performance improving with age and both ensembles exhibiting similar behavior under sum-based aggregation. Predictions for the Male class are consistently more stable, while those for the Female class display greater variability, particularly in the youngest age group. This difference becomes less pronounced in older strata, where both classes are recognized with higher accuracy. Importantly, although Male predictions tend to achieve higher recall, this does not imply that Female predictions are intrinsically more reliable. Rather, the results suggest that Female classifications may require stronger or more distinctive evidence, leading to fewer correct detections overall. Consequently, these patterns should be interpreted as reflecting differences in class-specific model sensitivity rather than straightforward differences in prediction confidence.

%\begin{enumerate}
%	\item \textbf{Model Capacity}: EfficientNet-B3 typically possesses a higher parameter count and depth than the ``Small'' (S) variant of MobileViT, potentially allowing it to capture more intricate morphological nuances.
%	\item \textbf{Architectural Inductive Bias}: As a pure Convolutional Neural Network (CNN), EfficientNet-B3 may have a stronger inductive bias for local spatial patterns in silhouettes, whereas the hybrid transformer-based MobileViT-S might require a larger dataset to fully leverage its global attention mechanisms.
%	\item \textbf{Optimization Sensitivity}: MobileViT architectures can be more sensitive to specific hyperparameter settings or the resolution of the input data compared to the highly optimized and robust EfficientNet framework.
%\end{enumerate}

\subsection{Prehistoric data: model application and interpretability on cave hand stencils}

This section presents the application of the inference and analysis pipeline described in~\ref{subsec:methodology} to the Paleolithic hand stencils. It consists of three parts, each reporting results obtained by aggregating evidence across all silhouette variants and model instances:\\[-3mm]

\noindent\textbf{Sex prediction results on prehistoric data}. The two ensembles, each comprising 10 model instances of EfficientNet-B3 and MobileViT-S trained on contemporary hand silhouettes from the RSNA Bone Age Challenge dataset, are applied to the 12 silhouette representations of the nine Paleolithic hand stencils. The ensemble strategy is explained below.\\[-3mm]

\noindent\textbf{Post-hoc latent-space analysis and manifold mapping}. Both contemporary and Paleolithic hand silhouettes are projected into 2D manifolds derived from EfficientNet-B3 latent spaces using UMAP. Each silhouette variant is classified via a $k$-NN rule based on the distribution of projected contemporary training samples. Predictions are aggregated across variants and EfficientNet-B3 instances to provide a complementary sex classification for each stencil.\\[-3mm]

\noindent\textbf{Post-hoc interpretability analysis via spatial attribution}. For each prehistoric hand image, visual attribution maps are generated using LayerCAM across the 12 silhouette variants and the 10 Efficient\-Net-B3 instances, yielding 120 maps per image. These maps are aggregated into a single, robust consensus attribution map for each cave image, highlighting spatial patterns relevant to sex prediction.

\subsubsection{Sex prediction results on prehistoric data}
\label{subsubsec:results-on-prehistoric-data}

\vspace{0.3cm}
This section provides the necessary details to understand the ensemble strategies and presents and discusses the aggregated results for sex prediction of the hand stencils.

Section~\ref{subsubsec:phase2} introduced the general methodology for sex prediction via score aggregation. In summary, two ensembles of 10 model instances each process the 12 silhouette variants of every hand stencil, yielding 120 predictions per ensemble. For each architecture, these predictions are aggregated using both sum- and max-based strategies to produce robust predictions. Each prediction is expressed as a probability distribution over two classes, Female and Male. %, i.e., two class probabilities that sum to one.

At the first level of aggregation, the outputs across the 10 ensemble instances are combined for each individual silhouette variant. Under sum aggregation, cumulative posterior probabilities are computed for each class, while under max aggregation, the maximum probability per class is selected. The consensus class for each silhouette variant is defined as the one with the highest aggregated score. Additionally, a classification margin is computed as the difference between the class ensemble scores (normalized for the sum rule). The sign of this margin indicates the ensemble's preferred class for that variant, while its magnitude, the ensemble's confidence.  

At the second aggregation level, the predictions obtained from the 12 silhouette variants are combined to determine the final classification of the original hand stencil. The final prediction is assigned to the majority class among the 12 variants and is evaluated using two confidence metrics: the Silhouette Support Rate (SSR) and the Average Classification Margin (ACM). The SSR quantifies the proportion of silhouette variants supporting the final consensus class. Given the two-class system without an abstention mechanism, the SSR ranges from 0.5 (maximum ambiguity) to 1 (complete consensus across all variants). The ACM is computed by first averaging the signed classification margins across all variants, and then taking the absolute value of that mean as a fine-grained confidence measure ranging from 0 (minimum certainty) to 1 (maximum confidence). 

Table~\ref{tab:sex_results} summarizes sex prediction results for the nine Paleolithic hand stencils using classifier ensembles based on two DNN architectures (EfficientNet-B3 and MobileViT-S), under both sum- and max-based fusion strategies. Final predictions are compared with prior or existing hypotheses derived from earlier statistical or morphometric approaches.

\begin{sidewaystable*}
\centering
\small
\caption{Sex attribution results for nine Upper Paleolithic hand stencils using EfficientNet-B3 and MobileViT-S ensembles, including fusion strategies, silhouette support rates (SSR), average classification margins (ACM), and comparison with prior archaeological hypotheses. F and M denote the Female and Male classes, respectively.}
\label{tab:sex_results}

\setlength{\tabcolsep}{3pt}
\renewcommand{\arraystretch}{1.1}

\begin{tabular}{l@{\hskip 6mm}c@{\hskip 5mm}c@{\hskip 5mm}c@{\hskip 6mm}cccccccc}
\toprule
\multicolumn{12}{c}{\textbf{Cases from El Castillo cave (selected for their high contrast and well-defined edges)}}\\
\midrule
\multirow{3}{*}{Cave Image*} & \multicolumn{3}{c}{\multirow{1}{*}{Prior / Existing Hypotheses}} 
& \multicolumn{4}{c}{EfficientNet-B3} 
& \multicolumn{4}{c}{MobileVit-S} \\
\cmidrule(lr){5-8}\cmidrule(lr){9-12}
& \multicolumn{3}{c}{\parencite{snow:2006}} 
& \multicolumn{2}{c}{Sum aggregation}
& \multicolumn{2}{c}{Max aggregation}
& \multicolumn{2}{c}{Sum aggregation}
& \multicolumn{2}{c}{Max aggregation} \\
\cmidrule(lr){2-4}\cmidrule(lr){5-6}\cmidrule(lr){7-8}\cmidrule(lr){9-10}\cmidrule(lr){11-12}
 & Step 1 & Step 2 & Inference
 & Class & SSR/ACM
 & Class & SSR/ACM
 & Class & SSR/ACM
 & Class & SSR/ACM \\
\midrule
img\_26 & Strong Female & Strong Female & Adult Female & F & 1/0.66 & F & 1/0.56 & F & 1/0.71 & F & 1/0.55 \\
img\_28 & Weak Female & Weak Male & Adolescent Male & F & 1/0.71 & F & 1/0.63 & F & 1/0.77 & F & 1/0.57 \\
img\_31 & Strong Female & Weak Male & Adolescent Male & M & 0.83/0.18 & M & 0.92/0.19 & F & 0.83/0.05 & F & 0.92/0.05 \\
img\_38 & Strong Female & Female & Adult Female & M & 1/0.26 & M & 1/0.32 & M & 1/0.24 & M & 1/0.24 \\
\bottomrule
\end{tabular}

\vspace{0.5em}
\footnotesize
\begin{tabular}{p{0.9\textheight}}
* Case names are consistent with the ID numbers shown in the original images.\\
%$^+$ Proportions of silhouette variants that support the predicted consensus class (silhouette support rate).
\end{tabular}

\vspace{1em}
\small
\begin{tabular}{lccccccccccc}
\toprule
\multicolumn{12}{c}{\textbf{Cases from the work \parencite{mollineda:2025}}}\\
\midrule
\multirow{3}{*}{Cave Image*} & \multicolumn{3}{c}{\multirow{2}{*}{Prior / Existing Hypotheses}} 
& \multicolumn{4}{c}{EfficientNet-B3} 
& \multicolumn{4}{c}{MobileVit-S} \\
\cmidrule(lr){5-8}\cmidrule(lr){9-12}
& & & 
& \multicolumn{2}{c}{Sum aggregation}
& \multicolumn{2}{c}{Max aggregation}
& \multicolumn{2}{c}{Sum aggregation}
& \multicolumn{2}{c}{Max aggregation} \\
\cmidrule(lr){2-4}\cmidrule(lr){5-6}\cmidrule(lr){7-8}\cmidrule(lr){9-10}\cmidrule(lr){11-12}
 & \parencite{wang:2010} & \parencite{snow:2013} & \parencite{mollineda:2025}
 & Class & SSR/ACM
 & Class & SSR/ACM
 & Class & SSR/ACM
 & Class & SSR/ACM \\
\midrule
Cosquer & Female & N/A & Female & F & 1/0.75 & F & 1/0.72 & F & 1/0.81 & F & 1/0.66 \\
Chauvet & Male & N/A & Strong Male & F & 1/0.29 & F & 1/0.33 & F & 0.75/0.08 & M & 0.58/0.04 \\
Unknown & Male & N/A & Strong Male & M & 0.83/0.09 & M & 0.67/0.04 & M & 0.92/0.17 & F & 0.58/0.03 \\
Pech Merle & Male & Female & Weak Male & M & 0.92/0.15 & M & 0.50/0.05 & F & 0.83/0.17 & F & 0.83/0.13 \\
El Castillo 25 & Male & Male & Strong Male & M & 1/0.68 & M & 1/0.62 & M & 1/0.79 & M & 1/0.75 \\
\bottomrule
\end{tabular}

\vspace{0.5em}
\footnotesize
\begin{tabular}{p{0.9\textheight}}
* Case names are consistent with those used in \parencite{mollineda:2025}.
\end{tabular}
\end{sidewaystable*}

The results indicate that ensembles produce stable and coherent predictions across architectures and aggregation strategies, with most cases showing agreement between sum and max rules within the same model. This internal consistency suggests that the aggregation process effectively reduces variability across the 120 individual predictions per stencil, yielding robust consensus outputs. In particular, cases with high SSR (i.e., $SSR\approx1$) and high ACM (i.e., $ACM\ge0.5$) can be interpreted as high-confidence predictions, reflecting strong agreement across both silhouette variants and ensemble members. This level of certainty is exemplified by the samples img\_26 (Female), img\_28 (Female), Cosquer cave (Female), and El Castillo 25 (Male), all of which exhibit complete agreement among the 12 silhouette variants ($SSR=1$) across both deep learning ensembles and both aggregation strategies, while also demonstrating strong model confidence ($ACM>0.5$).

A broader pattern emerges when comparing the two DNN architectures: EfficientNet-B3 typically yields more stable predictions than Mo\-bi\-le\-ViT-S, frequently accompanied by higher confidence measures. This trend is consistent with earlier observations on contemporary data, where EfficientNet-B3 demonstrated superior average performance, suggesting that its latent representations are more effective at capturing discriminative features for this task. Nonetheless, Mo\-bi\-le\-ViT-S maintains a high level of classification consistency, further validating the overall robustness and cross-model reliability of the ensemble approach. This evidence further supports the decision to conduct the post-hoc analyses using the EfficientNet-B3 ensemble.

When comparing the ensemble predictions with prior hypotheses, the results reveal a heterogeneous pattern of agreement and divergence. Several cases are consistent with earlier morphometric or statistical interpretations, particularly those historically regarded as less ambiguous. For example, img\_26, and Cosquer were unanimously classified as Female with high SSR and ACM values, aligning perfectly with existing archaeological hypotheses. Similarly, El Castillo 25 was strongly classified as Male across all configurations, in complete agreement with prior estimations. In contrast, strong divergences were identified in two other El Castillo cases (img\_28 and img\_38), where all ensembles produced highly consistent and confident predictions that differed from prior hypotheses. Cases like img\_31 and Pech Merle remain complex, as the models split between Male and Female predictions, mirroring the lack of consensus found in traditional morphometric or statistical approaches. 

Confidence measures (SSR and ACM) play a key role in interpreting these results. Cases with lower SSR or ACM values tend to correspond to historically ambiguous or controversial stencils, suggesting that the ensemble appropriately reflects underlying uncertainty rather than forcing overconfident decisions. Conversely, high-confidence cases indicate strong structural consistency across silhouette variants and model instances.

%These discrepancies highlight the potential of data-driven approaches to provide complementary perspectives, especially in cases where traditional methods yielded conflicting or uncertain conclusions.

%The remaining four cases (img\_31, Chauvet, Unknown, and Pech Merle) exhibited more uncertain and inconsistent profiles, characterized not only by discrepancies between prior hypotheses but also by disagreements between ensembles and low confidence measures. From a more qualitative perspective, both EfficientNet-B3-based ensembles aligned with the most relevant prior interpretations in 6 out of the 9 Paleolithic cases, whereas the two MobileViT-S-based ensembles were consistent with only 3 out of 9 cases.

%In this regard, cases with low support values were more likely to show either architectural disagreement or prior interpretive ambiguity, proving their usefulness for assessing prediction reliability. 

Overall, the results suggest that while AI can provide highly confident sex attributions for well-defined stencils, cases with low SSR and ACM values indicate inherent morphological ambiguity in the prehistoric data. The combination of aggregation strategies and confidence metrics provides a nuanced view of the results, allowing both agreement and uncertainty to be explicitly quantified.

%Overall, Table 4 supports the conclusion that the proposed ensemble-based framework yields robust, interpretable, and uncertainty-aware predictions, while also revealing meaningful differences with prior approaches. 

\subsubsection{Post-hoc latent-space analysis and manifold mapping}

\vspace{0.3cm}

\begin{table*}[t]
\centering
\caption{Aggregated $k$-NN classification performance within 2D manifolds across the 12 silhouette variants and the 10 EfficientNet-B3 instances.}
\label{tab:umap_results}

\begin{tabular}{lcccc}
\hline
& \multicolumn{2}{c}{$k$-NN} & \multicolumn{2}{c}{EfficientNet-B3} \\
\hline
Cave Image & Predicted Class & SSR$^+$ & Predicted Class & SSR/ACM$^*$ \\
\hline
img\_26 & Female & 0.982 & Female & 1/0.61 \\
img\_28 & Female & 0.966 & Female & 1/0.67 \\
img\_31 & Male & 0.697 & Male & 0.87/0.18 \\
img\_38 & Male & 0.821 & Male & 1/0.29 \\
Cosquer & Female & 0.989 & Female & 1/0.73 \\
Chauvet & Female & 0.656 & Female & 1/0.31 \\
Unknown & Male & 0.793 & Male & 0.75/0.06 \\
Pech Merle  & Male & 0.823 & Male & 0.71/0.10 \\
El Castillo 25 & Male & 0.981 & Male & 1/0.65 \\
\hline
\end{tabular}

\vspace{0.5em}
\footnotesize
\centering
\begin{tabular}{c} % p{0.75\textheight}
$^+$ Average of the mean SSRs (computed across all 2D manifolds) over the interval $k \in [20, 50)$.\\
%$^+$ Mean SSR averaged over the range $k \in [10, 50]$, where each $k$-specific value represents the average SSR across all 2D manifold instances.\\
* Average SSR and ACM values across both fusion strategies for the EfficientNet-B3 ensemble (derived from Table~\ref{tab:sex_results}).\\
\end{tabular}
\end{table*}

%In search of additional evidence, 
Each projected silhouette variant is classified using the $k$-NN rule based on the distribution of contemporary training samples embedded within 2D manifolds. %derived from the latent spaces of EfficientNet-B3 models. 
The resulting predictions are aggregated, first across silhouette variants and then across 2D mappings, yielding a complementary sex classification hypothesis for each stencil. Consistent with the confidence measures defined in~\ref{subsubsec:results-on-prehistoric-data}, the SSR is computed for each hand stencil within each 2D manifold, and the resulting SSR values are averaged across the ten mappings. Accordingly, the SSR ranges from 0.5, indicating maximum ambiguity, to 1, indicating complete consensus across all silhouette variants.

\begin{figure}[t]
	\centering
	\includegraphics[width=\linewidth]{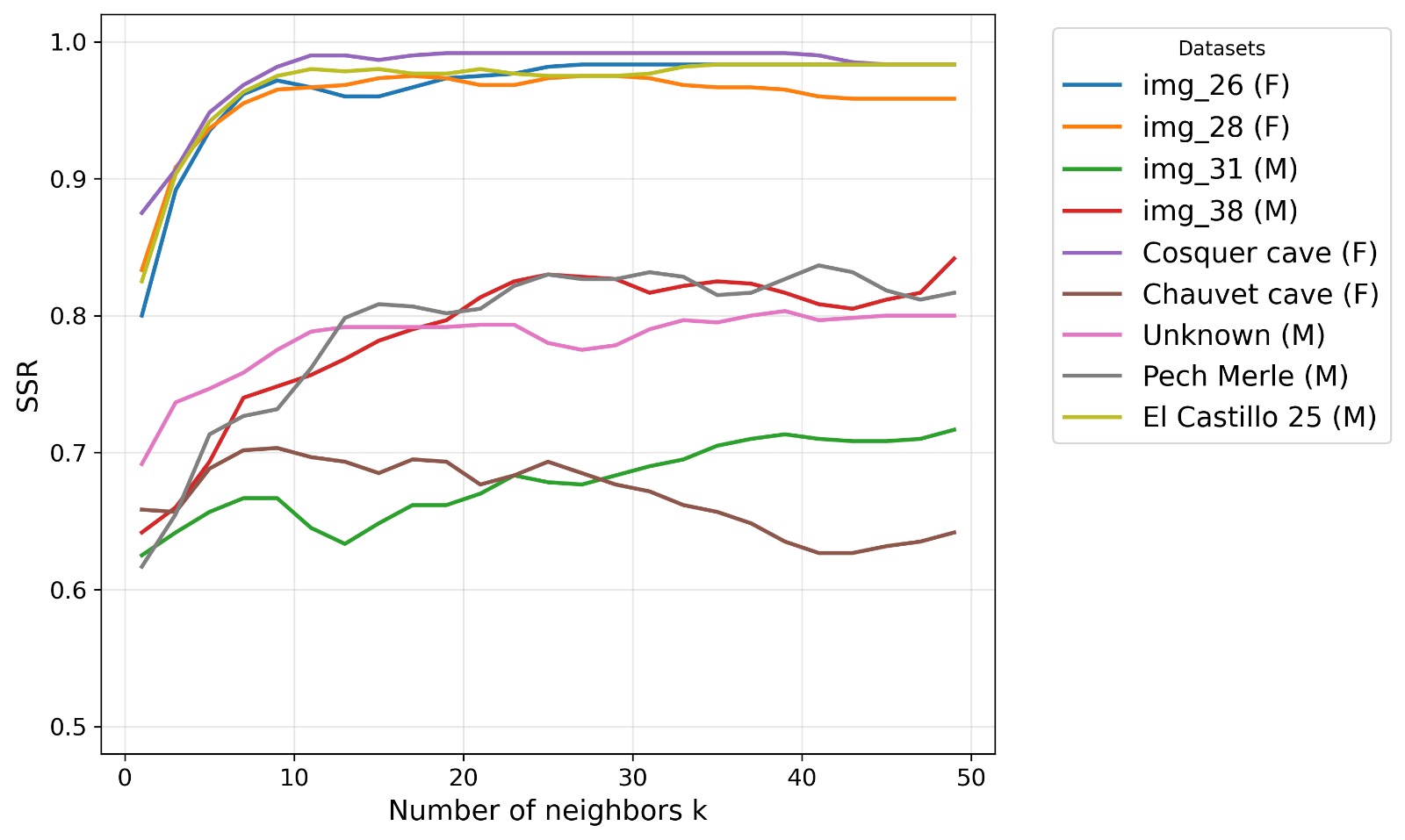}
	\caption{Mean SSR (support for the predicted class across all silhouette variants) as a function of $k$. A sensitivity analysis was conducted over the range $k \in [1, 50)$. Since SSR remains consistently above 0.5, the $k$-NN predictions are shown to be robust and largely insensitive to the choice of $k$.}
	\label{fig:SSR-as-func-of-k}
\end{figure}

Rather than selecting an arbitrary value for $k$, a sensitivity analysis was conducted across the range $k \in [1, 50)$, leveraging the high density of the projected contemporary data. For each value of $k$, the mean SSR was calculated across all 2D mappings to generate a curve representing SSR behavior as a function of $k$ for each hand stencil (see Figure~\ref{fig:SSR-as-func-of-k}). This analysis revealed that the $k$-NN aggregation predictions are invariant to changes in $k$ (SSR initially scales with $k$ before stabilizing significantly above 0.5), de\-monstrating the robustness of the class-conditional densities within the mappings and aggregation approach based on $k$-NN. As illustrated in Figure~\ref{fig:SSR-as-func-of-k}, the SSR curves stabilize for $k \geq 20$; then, the average SSR across the interval $[20, 50)$ was adopted as the final confidence measure for the $k$-NN classification of each stencil.

Table~\ref{tab:umap_results} summarize the sex classification results for the nine cave hand stencils, including both the predictions obtained using the $k$-NN rule on the 2D manifold mappings and those produced by the EfficientNet-B3 ensembles. This comparison aims to evaluate the degree of convergence between the highly interpretable 2D representations and the DNN models, which exhibit strong generalization capabilities.

Overall, there is strong concordance between the $k$-NN classifications and the EfficientNet-B3 ensemble predictions. In all nine cases, the predicted sex label obtained from the $k$-NN classifier matches the label reported by the ensemble. This consistency suggests that the class structure encoded in the high-dimensional latent spaces is largely preserved in the reduced two-dimensional manifold and remains separable using a simple distance-based method.

Four cases show particularly strong support values across both averaging schemes (embedding-space averaging and fusion-strategy averaging), indicating robust local clustering in the 2D space. These include img\_26 and img\_28 from El Castillo Cave (Female, 0.966–1.0 support), the Cosquer case from Cosquer Cave (Female, 0.989–1.0 support), and El Castillo 25 (Male, 0.981–1.0 support). In these cases, the k-NN classifier identifies highly discriminant neighborhoods in the 2D projection, consistent with the decisive ensemble predictions. The classification for these four hand stencils is illustrated in Figure~\ref{fig:umap}. The three points (silhouettes) corresponding to each case are positioned consistently outside the region of confusion between the two classes.

Moderate support values are observed in img\_31 (Male, 0.697–0.87), img\_38 (Male, 0.821–1.0), the Chauvet case (Female, 0.656–1.0), the unknown case (Male, 0.793–0.75), and the Pech Merle case (Male, 0.823–0.71). Although their support levels are lower than in the most robust examples, the predicted class remains stable across both approaches. The reduced support values indicate less compact clustering of the 12 silhouette variants in the 2D space, but do not result in label inversion relative to the ensemble output. In contrast to the higher-confidence cases, as illustrated in Figure~\ref{fig:umap}, the points associated with these five hands are located much closer to the inter-class overlap region, thereby explaining the lack of consensus among the different classification rules.

Notably, no discrepancies were observed between the $k$-NN classifier applied to the 2D manifolds and the EfficientNet-B3 ensemble predictions. While modest variations in support metrics occurred, they did not alter the final class assignments. This alignment suggests that the sex-specific structure captured within the high-dimensional latent representations learned by EfficientNet-B3 instances is robust enough to remain discriminable even after aggressive unsupervised dimensionality reduction and to be recoverable through a simple neighborhood-based decision rule.

%In summary, the $k$-NN results in the 2D UMAP space provide complementary and convergent evidence for the ensemble-based sex classifications. The consistent label agreement across all nine hand stencils indicates that the latent feature space learned by EfficientNet contains stable class structure that is preserved under dimensionality reduction and recoverable through a simple neighborhood-based decision rule.

\subsubsection{Post-hoc interpretability analysis via spatial attribution}
\label{subsect:results-model-interpretability}

\vspace{0.3cm}

\begin{figure}[t]
    \centering
    \includegraphics[width=\linewidth]{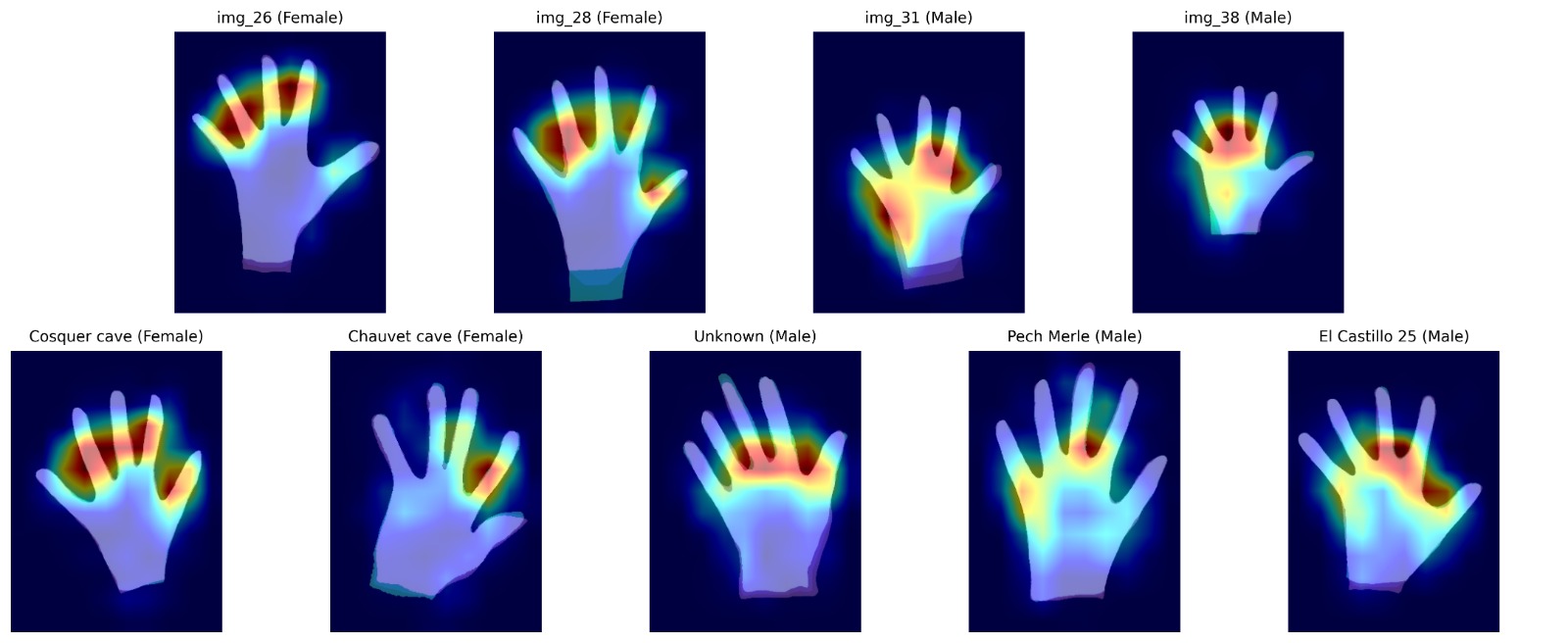}
    \caption{Robust LayerCAM attribution maps obtained by geometric median aggregation across 12 silhouette variants and 10 EfficientNet model instances.}
    \label{fig:attmaps}
\end{figure}

\begin{sidewaystable*}[htbp]
\centering
\caption{Case-by-case interpretive analysis integrating spatial attribution patterns and aggregation model confidence measures.}
\label{tab:interpretability_summary}
\renewcommand{\arraystretch}{1.2}
\begin{tabular}{l l >{\raggedright\arraybackslash}p{5.2cm} c c c >{\raggedright\arraybackslash}p{5.2cm}}
\hline
\textbf{Stencil} & \textbf{Sex} & \textbf{Attribution Pattern} & \textbf{Spatial} & \textbf{Ensemble} & \textbf{k-NN} & \textbf{Interpretive Assessment} \\
 &  &  & \textbf{Coherence} & \textbf{SSR} & \textbf{SSR} &  \\
\hline

img\_26 & F &
Broad activation across the index, middle and ring fingers with interdigital emphasis. &
High & 1.0 & High &
Morphologically stable female pattern consistent with maximal ensemble agreement. \\
%high predictive and manifold stability. \\
%Morphologically stable female pattern with strong latent-space clustering and maximal ensemble agreement. \\

img\_28 & F &
Focal activation in the central ring finger with interdigital emphasis. &
High & 1.0 & High &
Structurally compact female pattern consistent with maximal ensemble agreement. \\

img\_31 & M &
%Activation is somewhat dispersed, primarily across the proximal phalanges and the metacarpophalangeal joints of the little and index fingers.
Dispersed activation mainly in the proximal regions of the little and index fingers. &
Moderate & 0.83--0.92 & Low &
A dispersed activation pattern reflects architectural disagreement and small classification margins. \\

img\_38 & M &
Sharp, focal activation in proximal middle-ring regions. &
High & 1.0 & Moderate &
Internally confident classification despite reversal of prior interpretation. \\

Cosquer & F &
%Broad central finger relevance and interdigital sensitivity. &
Broad activation across the index, middle and ring fingers with interdigital sensitivity. &
High & 1.0 & High &
Strong agreement across models, silhouette variants and prior hypotheses; stable embedded representation. \\

Chauvet & F &
Diffuse activation in the mid region of the index fingers. &
Low & 0.58--1 & Moderate &
Structurally ambiguous female case mirrors discrepancy with prior hypotheses and architectural disagreement. \\

Unknown & M &
Sharp, focal activation at palm–finger junction. &
Moderate & 0.58--0.92 & Moderate &
Stable but not compact; moderate support across decision approaches. \\

Pech Merle & M &
Weakly focused activation at palm–finger junction. &
Moderate & 0.50--0.92 & Moderate &
Morphologically ambiguous case; mirrors historical interpretive discrepancy and architectural disagreement. \\

El Castillo 25 & M &
Focal activation at palm–finger junction. &
% Highly focal proximal finger activation with minimal peripheral dispersion. &
High & 1.0 & High &
Strongly stable male pattern with maximal ensemble and latent-space agreement. \\

\hline
\end{tabular}
\end{sidewaystable*}

To assess whether the predicted sex labels were supported by coherent and anatomically meaningful visual cues, 120 LayerCAM attribution maps (12 silhouette variants $\times$ 10 EfficientNet-B3 instances) were generated for each stencil and aggregated using the geometric median. This aggregation suppresses contour-specific artifacts and model stochasticity, yielding a robust prototype explanation per hand. The nine resulting maps, which are shown in Fig.~\ref{fig:attmaps}, represent stable spatial patterns across both silhouette uncertainty and model replication.

Across all nine cases, relevance is consistently concentrated along the fingers and interdigital spaces, particularly around middle and proximal phalanges and the metacarpal heads, while the central palm and wrist regions show minimal activation. This spatial selectivity suggests that the models rely primarily on finger geometry and relative spacing (features consistent with sexually dimorphic morphology) rather than on global hand size or background structure. Table~\ref{tab:interpretability_summary} brings together predictive, latent-space, and interpretability indicators for each case. 

%\paragraph{Female-classified stencils.}
A clear distinction emerges within the four female predictions. A first group composed of img\_26, img\_28, and the Cosquer case shares a highly consistent relevance structure characterized by distributed activation across the central phalangeal regions of the index, middle and ring fingers, with consistent emphasis on interdigital spacing and some minor activation on the thumb. These maps are spatially compact and anatomically organized. Importantly, all three cases exhibit maximal or near-maximal silhouette support in Table~\ref{tab:sex_results} and strong $k$-NN support in the 2D manifolds in Table~\ref{tab:umap_results}, indicating both predictive stability and compact latent-space clustering. In contrast, the Chauvet stencil exhibits a more diffuse attribution pattern, irregularly distributed across the middle and proximal phalanges of the index and middle fingers. %, with only weak activation near the first metacarpal. 
This case corresponds to the lowest $k$-NN support among female classifications in Table~\ref{tab:umap_results} and to one of the largest discrepancies between prior archaeological hypotheses and deep learning predictions (Table~\ref{tab:sex_results}). The spatial dispersion observed in the Chauvet map therefore aligns with its reduced structural support and interpretive ambiguity.

%\paragraph{Male-classified stencils.}
The five male-classified cases generally exhibit less focal patterns, with strong activation around the me\-ta\-car\-pophalangeal joints of fingers, and occasionally extending %, with lower intensity, 
toward the ulnar side of the palm below the little finger. 
%around the bases of the index and middle fingers and the metacarpal region. 
Among them, img\_38 and El Castillo 25 (both with maximal support in Table~\ref{tab:sex_results}) present %sharply localized and 
well-defined spatial clusters. Their compact attribution patterns mirror their strong ensemble agreement and high latent-space support in Table~\ref{tab:umap_results}. In contrast, img\_31 and Pech Merle show broader activation zones consistent with intermediate or reduced support values (Table~\ref{tab:sex_results}). Notably, Pech Merle, characterized by interpretive disagreement, exhibits one of the most diffuse patterns among male cases, paralleling its lower silhouette support and architectural divergence. As in the Chauvet example, spatial dispersion appears to reflect underlying morphological or representational ambiguity rather than classification instability alone.

%\paragraph{Attribution stability and predictive support.}
%A qualitative correspondence is observed between silhouette support rates (Table~\ref{tab:sex_results}), latent-space separability (Table~\ref{tab:umap_results}), and spatial concentration of attribution maps. Cases with maximal internal agreement (support $\approx 1$) consistently display compact and anatomically coherent activation patterns, whereas cases with lower support or architectural disagreement show more diffuse relevance distributions. Because each map aggregates variability across contour realizations and independent model instances, these spatial patterns can be interpreted as stable morphological signals. In this sense, geometric median LayerCAM aggregation functions not only as an interpretability tool but also as a complementary diagnostic of epistemic stability within the broader uncertainty-aware framework.

\section{Discussion}
\label{sec:discussion}

The proposed methodology was structured as a multi-layered uncertainty management system. Instead of treating uncertainty as noise to be suppressed, it is modeled and propagated across the full analytical pipeline before being resolved through hierarchical aggregation. Uncertainty is addressed at the source level (dual image processing), annotation level (dual contour extraction), morphological level (structured silhouette perturbations), representational level (multiple three-channel compositions), model level (architectural diversity and stochastic replication), and decision level (ensemble fusion with silhouette support rates). Each stage introduces controlled variability prior to aggregation, ensuring that final predictions emerge from the convergence of multiple imperfect but systematically structured observations.

The decision-making process incorporates criteria beyond simple probability estimates. Structural validation was conducted using an unsupervised UMAP projection followed by non-parametric $k$-NN classification to test whether class separability persists under aggressive dimensionality reduction. Additionally, spatial attribution maps are aggregated via the geometric median to ensure explanation stability across contour variants and model instances. The interpretability results provide independent support for the internal coherence of the framework. Because each LayerCAM map aggregates 120 individual attribution instances, the resulting spatial patterns reflect relevance structures that are stable under both contour uncertainty and stochastic model variation. 

Across cases, activation patterns consistently target anatomically meaningful regions of the hand, particularly the fingers, interdigital spaces, proximal and middle phalanges, and metacarpal heads, while assigning comparatively little relevance to the palm and wrist. This observation is noteworthy because several of these structures have previously been identified as important sources of sexual dimorphism in skeletal studies. For example, the proximal phalanges have been reported to exhibit significant sex-related differences, with the thumb showing particularly pronounced dimorphism. The correspondence between these anthropologically established markers and the regions highlighted by the attribution maps suggests that the classifiers are relying on biologically plausible morphological cues rather than on spurious image characteristics or global size-related proxies. %Furthermore, the recurrent emphasis on finger geometry and relative spacing across diverse stencil representations and model instances indicates that these features constitute a stable source of discriminative information within the learned representations. 
Although attribution maps cannot establish causal relationships between anatomical traits and classification outcomes, the observed spatial consistency provides additional evidence that the models capture meaningful aspects of hand morphology associated with sexual differentiation. The convergence of these activation patterns across multiple silhouette variants and ensemble members further suggests that the identified anatomical regions are robust to contour uncertainty and model stochasticity, thereby strengthening confidence in the interpretability of the resulting predictions.

Moreover, a qualitative correspondence emerges between attribution coherence, ensemble confidence rates, and latent-space separability. Cases exhibiting maximal internal consensus ($SSR \approx 1$) display sharply localized and anatomically organized activation patterns. By contrast, cases associated with architectural disagreement or reduced silhouette support show more diffuse and spatially heterogeneous relevance distributions. This alignment indicates that spatial dispersion in the geometric median maps may function as a visual correlate of epistemic ambiguity rather than mere saliency noise. The layered design thus transforms uncertainty from a liability into an analyzable signal. Agreement across predictive aggregation, latent-space validation, and attribution stability provides convergent evidence for morphologically stable classifications, whereas divergence across these layers highlights structurally ambiguous cases. The framework therefore functions not merely as a predictive system, but as a structured mechanism for diagnosing stability in the absence of archaeological ground truth.

From a practical perspective, the proposed framework is intended to support archaeological interpretation through the joint analysis of predictions, confidence measures, latent-space organization, and attribution patterns, rather than through binary sex labels alone. Hand stencils exhibiting strong consistency across ensemble configurations, high silhouette support rates, large classification margins, consistent positioning within the latent-space manifolds, and agreement between ensemble and k-NN classifications may be considered relatively robust candidates for sex attribution. In contrast, cases characterized by reduced support rates, small classification margins, proximity to the overlap region between classes in the latent-space projections, or disagreement among analytical components should be interpreted as inherently ambiguous. Such cases are not necessarily classification failures; instead, they may reflect genuinely intermediate morphologies, insufficient visual information, or limitations of the available reference population. Accordingly, the confidence indicators produced by the framework should be treated as archaeological evidence in their own right. High-confidence predictions can contribute to broader demographic interpretations, whereas low-confidence cases may be more appropriately reported as indeterminate rather than being forced into categorical assignments. This perspective shifts the focus from obtaining definitive classifications toward evaluating the strength and reliability of the available evidence, promoting more transparent and reproducible interpretations of prehistoric hand stencils.

A fundamental limitation of this study arises from the difficulty of establishing how closely contemporary reference populations resemble the Upper Paleolithic populations to which the framework is ultimately applied. Given the considerable temporal and evolutionary separation between these groups, substantial differences in the distribution of sex-related hand morphology are plausible, yet their magnitude remains unknown. 
%A fundamental limitation of this study arises from the unavoidable discrepancy between the contemporary populations used for model development and the Upper Paleolithic populations to which the framework is ultimately applied. 
All supervised learning approaches for prehistoric hand stencil analysis, whether based on morphometric measurements, engineered features, or deep learning, necessarily rely on modern reference data because no ground-truth sex labels exist for archaeological hand stencils. Consequently, the models assume that at least part of the morphological signatures of sexual dimorphism observed in present-day human populations remain sufficiently stable across long temporal scales. While this assumption is supported by the persistence of broad anatomical differences between sexes, it cannot be directly verified for Upper Paleolithic populations. Evolutionary, demographic, nutritional, developmental, and population-specific factors may have influenced hand morphology over thousands of years, potentially altering the distribution of sex-related traits. Previous studies have already demonstrated that predictive functions developed in one contemporary population may generalize poorly to another due to differences in average hand morphology and population structure. Therefore, even highly consistent predictions should not be interpreted as direct determinations of biological sex, but rather as probabilistic inferences relative to the morphological patterns learned from the available contemporary reference population. Although future methodologies based on unlabeled target-domain data may help narrow the gap between contemporary and prehistoric populations, the feasibility of such approa\-ches remains strongly conditioned by the limited number, quality, and representativeness of currently available archaeological samples. Consequently, the proposed framework should be understood as a tool for generating evidence-based hypotheses and quantifying their internal consistency, rather than as a mechanism for obtaining definitive sex attributions.

This framework could be further advanced by enhancing its ability to represent input variability and quantify uncertainty within the predictive process. Moving toward a more continuous and expressive representation of plausible hand shapes would enable a richer characterization of morphological variability and a more robust assessment of prediction stability under structured perturbations. In parallel, incorporating mechanisms to capture uncertainty in model predictions would shift from point estimates to predictive distributions, enabling a more systematic quantification of confidence and its integration into result interpretation.

More broadly, virtually all stages of the analytical pipeline offer opportunities for systematic scaling to improve uncertainty modeling. From early image processing and manual contour delineation—where independent realizations can be incorporated—each stage can be expanded to propagate variability forward. The cumulative combination of outputs across stages enables a more comprehensive exploration of plausible interpretations, strengthening the robustness of final predictions. Together, these directions point toward a more fully probabilistic formulation of the framework, enhancing its ability to disentangle data-driven ambiguity from model uncertainty, particularly in the absence of reliable ground truth.

\section{Conclusions}
\label{sec:conclusions}

This study introduced a multi-layered uncertainty-aware framework for sex attribution in prehistoric hand stencils. Rather than treating uncertainty as an undesirable artifact to be eliminated, the proposed methodology models, propagates, and aggregates uncertainty across multiple levels of analysis, including image acquisition, manual annotation, silhouette representation, model training, and decision making. The resulting framework combines controlled variability with hierarchical aggregation, allowing final classifications to emerge from the convergence of multiple complementary observations rather than from a single deterministic prediction.

Experimental results on contemporary hand silhouettes proved that ensemble-based deep learning models can reliably capture sexually dimorphic morphological patterns, particularly in age groups where such traits are more strongly expressed. When transferred to prehistoric hand stencils, the framework produced stable predictions for several cases while simultaneously identifying others as intrinsically ambiguous through reduced silhouette support and lower classification margins. These confidence indicators offer valuable guidance in mitigating the overinterpretation of uncertain archaeological evidence.

A key contribution of the study is the triangulation of evidence through three complementary perspectives: ensemble-based classification, latent-space validation through unsupervised learning of inter\-pre\-table 2D manifold, and aggregated LayerCAM explanations. The strong agreement observed among these components indicates that the inferred classifications are supported by consistent latent representations and anatomically meaningful visual cues. Conversely, cases exhibiting disagreement across these analyses reveal regions of epistemic uncertainty that warrant cautious interpretation.

More broadly, the proposed framework demonstrates that uncertainty itself can be treated as an informative signal. Agreement across predictive, structural, and interpretability analyses provides convergent evidence for morphologically stable classifications, whereas divergence highlights cases where the available evidence remains insufficient for definitive attribution. In this sense, the methodology functions not only as a predictive system but also as a tool for assessing the reliability of archaeological inferences in the absence of ground truth.

Future work should move the framework toward more explicitly probabilistic formulations, strengthening its capacity to represent input variability and to quantify uncertainty in predictions. These developments may help disentangle morphological ambiguity from model-related uncertainty.

\section*{Acknowledgments}

The authors would like to express their sincere gratitude to Professor Roberto Ontañón Peredo for kindly providing the scaled images of the El Castillo cave hand stencils used in Figures~\ref{fig:intro_four_hands} and~\ref{fig:five_hands}, as well as in part of Figure~\ref{fig:contouring}. These images were instrumental in both the development of the proposed methodology and the presentation of the results.

The authors also acknowledge the partial financial support provided by grant AIA2025-163919-C54, funded by MICIU/AEI/10.13039/501100011033 (Spain).

%The authors would like to express their sincere gratitude to Professor Roberto Ontañón Peredo for kindly providing the scaled images of the El Castillo cave hand stencils used in Figures~\ref{fig:intro_four_hands} and~\ref{fig:five_hands}, and partially in Figure~\ref{fig:contouring}. These materials were instrumental in supporting both the methodological development and the presentation of the results. The authors would also like to acknowledge that this work was partially supported by grant AIA2025-163919-C54, funded by MICIU/AEI/10.13039/50110001\-1033 (Spain).

%\section*{CRediT authorship contribution statement}

%\textbf{Karel Becerra}: Conceptualization, Data curation, Investigation, Resources, Software, Validation, Visualization, Writing – review and editing. \textbf{Boris Mederos}: Conceptualization, Formal analysis, Investigation, Methodology, Software, Validation, Visualization, Writing – review and editing. \textbf{Ramón A. Mollineda}: Conceptualization, Formal analysis, Investigation, Methodology, Software, Visualization, Writing – original draft, Writing – review and editing.

\section*{Declaration of competing interest}

The authors declare that there is no conflict of interest that could affect the independence of the research reported in this paper.

\section*{Data availability}

To promote transparency, reproducibility, and independent verification of the reported results, the data\-sets generated in this study are publicly available.\footnote{The data generated in this work is available through this link: \href{https://www.kaggle.com/datasets/karelbecerra/decoding-the-past-in-prehistoric-hand-stencils}{https://www.kaggle.com/datasets/karelbecerra/decoding-the-past-in-prehistoric-hand-stencils}}. The source code developed for the classification pipeline and the visualization of the experimental results is also publicly available.\footnote{The code written in this work is available through this link: \href{https://github.com/karelbecerra/hand-paintings-in-rock-art}{https://github.com/karelbecerra/hand-paintings-in-rock-art}}

%\section*{Funding sources}

%This research did not receive any specific grant from funding agencies in the public, commercial, or not-for-profit sectors.

\section*{Ethics approval}

Not applicable.

\section*{Declaration of generative AI and AI-assisted technologies in the manuscript preparation process}

During the preparation of this work, the authors used generative artificial intelligence tools, specifically ChatGPT (GPT-5.5) developed by OpenAI, to improve the readability and clarity of selected sections of the manuscript, and Gemini Notebook to generate infographics. All scientific ideas, analyses, and interpretations were conceived and developed independently by the authors. The authors carefully reviewed and edited all AI-assisted content and assume full responsibility for the accuracy, integrity, and conclusions of the published article.

\printbibliography

\end{document}